\documentclass[nohyperref]{article}

\usepackage{microtype}
\usepackage{multirow}
\usepackage{graphicx}
\usepackage{subcaption}
\usepackage{booktabs} 

\usepackage{hyperref}
\usepackage{color}
\usepackage{float} 
\usepackage{caption}
\usepackage{enumitem}

\usepackage{pifont}
\newcommand{\cmark}{\ding{51}}%
\newcommand{\xmark}{\ding{55}}%

\usepackage[accepted]{icml2022}

\usepackage{amsmath}
\usepackage{amssymb}
\usepackage{mathtools}
\usepackage{amsthm}

\usepackage{color, colortbl}
\definecolor{LightCyan}{rgb}{0.88,1,1}

\usepackage[capitalize,noabbrev]{cleveref}

\theoremstyle{plain}
\newtheorem{theorem}{Theorem}[section]
\newtheorem{proposition}[theorem]{Proposition}

\theoremstyle{definition}

\theoremstyle{remark}

\usepackage[textsize=tiny]{todonotes}

\newcommand{\x}{{\pmb{x}}}
\newcommand{\w}{\pmb{w}}
\newcommand{\s}{\pmb{s}}
\newcommand{\f}{\pmb{f}}
\newcommand{\g}{\pmb{g}}
\newcommand{\h}{\pmb{h}}
\newcommand{\z}{\pmb{z}}
\newcommand{\eye}{\pmb{I}}
\newcommand{\zero}{\pmb{0}}
\newcommand{\one}{\pmb{1}}
\newcommand{\eps}{\pmb{\epsilon}}
\newcommand{\bmu}{\pmb{\mu}}
\newcommand{\bSigma}{\pmb{\Sigma}}

\usepackage{etoolbox}
\newcommand{\zerodisplayskips}{%
	\setlength{\abovedisplayskip}{5pt}%
	\setlength{\belowdisplayskip}{5pt}%
	\setlength{\abovedisplayshortskip}{5pt}%
	\setlength{\belowdisplayshortskip}{5pt}}
\appto{\normalsize}{\zerodisplayskips}
\appto{\small}{\zerodisplayskips}
\appto{\footnotesize}{\zerodisplayskips}

\icmltitlerunning{Diffusion Models for Adversarial Purification}

\begin{document}
	
	\twocolumn[
	\icmltitle{Diffusion Models for Adversarial Purification}
	
	
	
	
	\begin{icmlauthorlist}
		\icmlauthor{Weili~Nie}{yyy}
		\icmlauthor{Brandon Guo}{comp}
		\icmlauthor{Yujia Huang}{comp}
		\icmlauthor{Chaowei Xiao}{yyy}
		\icmlauthor{Arash Vahdat}{yyy}
		\icmlauthor{Anima Anandkumar}{yyy,comp}
	\end{icmlauthorlist}
	
	\icmlaffiliation{yyy}{NVIDIA}
	\icmlaffiliation{comp}{Caltech}
	
	\icmlcorrespondingauthor{Weili Nie}{wnie@nvidia.com}
	
	\icmlkeywords{Machine Learning, diffusion models, adversarial robustness}
	
	\vskip 0.3in
	]
	
	
	
	\printAffiliationsAndNotice{}  

	\begin{abstract}
		
		Adversarial purification refers to a class of defense methods that remove adversarial perturbations using a generative model.
		These methods do not make assumptions on the form of attack and the classification model, and thus can defend pre-existing classifiers against unseen threats. However, their performance currently falls behind adversarial training methods.
		In this work, we propose 
		\textit{DiffPure}
		that uses diffusion models for adversarial purification: Given an adversarial example, we first diffuse it with a small amount of noise following a forward diffusion process, and then recover the clean image through a reverse generative process. 
		To evaluate our method against strong adaptive attacks in an efficient and scalable way,
		we propose to use the adjoint method to compute full gradients of the reverse generative process. 
		Extensive experiments on three image datasets including CIFAR-10, ImageNet and CelebA-HQ with three classifier architectures including ResNet, WideResNet and ViT demonstrate that our method achieves the state-of-the-art results, outperforming current adversarial training and adversarial purification methods, often by a large margin. 
		Project page: \url{https://diffpure.github.io}.
	\end{abstract}
	
	\vspace{-19pt}
	\section{Introduction}
	\label{intro}
	\vspace{-3pt}
	
	Neural networks are vulnerable to adversarial attacks: adding imperceptible perturbations to the input can mislead trained neural networks to predict incorrect classes~\cite{szegedy2014intriguing,goodfellow2015explaining}. There have been many works on 
	defending neural networks against such adversarial attacks~\cite{madry2018towards,song2018pixeldefend,gowal2020uncovering}. Among them, 
	\textit{adversarial training}~\cite{madry2018towards}, which trains neural networks on adversarial examples, has become a standard defense form, due to its effectiveness~\cite{zhang2019theoretically,gowal2021improving}. 
	However, most adversarial training methods can only defend against a specific attack that they are trained with.
	Recent works on defending against unseen threats add a carefully designed threat model into their adversarial training pipeline, but they suffer from a significant performance drop~\cite{laidlaw2021perceptual,dolatabadi2021ell_}.
	Additionally, the computational complexity of adversarial training is usually  higher than standard training~\cite{wong2020fast}.
	
	
	In contrast, another class of defense methods, often termed \textit{adversarial purification}~\cite{shi2020online,yoon2021adversarial}, relies on generative models to purify adversarially perturbed images before 
	classification~\cite{samangouei2018defense,hill2021stochastic}. 
	Compared to the adversarial training methods, adversarial purification can defend against unseen threats 
	in a plug-n-play manner without re-training the classifiers. This is because the generative purification models are trained independently from both threat models and classifiers.
	Despite these advantages, their performance usually falls behind current adversarial training methods~\citep{croce2020reliable}, in particular against adaptive attacks where the attacker has the full knowledge of the defense method~\cite{athalye2018obfuscated,tramer2020adaptive}.
	This is usually attributed to the shortcomings of current generative models that are used as a purification model, such as mode collapse in GANs~\citep{goodfellow2014generative}, low sample quality in energy-based models (EBMs)~\cite{lecun2006tutorial}, and the lack of proper
	randomness~\cite{pinot2020randomization}.
	
	Recently, diffusion models have emerged as powerful generative models~\cite{ho2020denoising,song2021score}. These models have demonstrated strong sample quality, beating GANs in image generation~\cite{dhariwal2021diffusion, vahdat2021score}. They have also exhibited strong mode coverage,
	indicated by high test likelihood~\cite{song2021maximum}. Diffusion models consist of two processes: (i) a forward diffusion process that converts data to noise by gradually adding noise to the input, and (ii) a reverse generative process that starts from noise and generates data by denoising one step at a time.
	Intuitively in the generative process, diffusion models purify noisy samples, playing a similar role of a purification model.
	Their good generation quality and diversity ensure the purified images closely follow the original distribution of clean data.
	Moreover, the stochasticity in diffusion models can make a powerful stochastic defense~\cite{he2019parametric}. 
	These properties make diffusion models an ideal candidate for generative adversarial purification.

	
	
	\textbf{We summarize our main contributions as follows:}
	\vspace{-8pt}
	\begin{itemize}[leftmargin=*]\setlength\itemsep{-0em}
		\item We propose \textit{DiffPure}, the first adversarial purification method that uses the forward and reverse processes of pre-trained diffusion models to purify adversarial images.
		\item We provide a theoretical analysis of the amount of noise added in the forward process such that it removes adversarial perturbations without destroying label semantics.
		\item We propose to use the adjoint method to efficiently compute full gradients of the reverse generative process in our method for evaluating against strong adaptive attacks.
		\item We perform extensive experiments to demonstrate that our method achieves the new start-of-the-art on various adaptive attack benchmarks.
		\vspace{-6pt}
	\end{itemize}
	
\begin{figure}[t]
		\centering
		\includegraphics[width=0.98\linewidth]{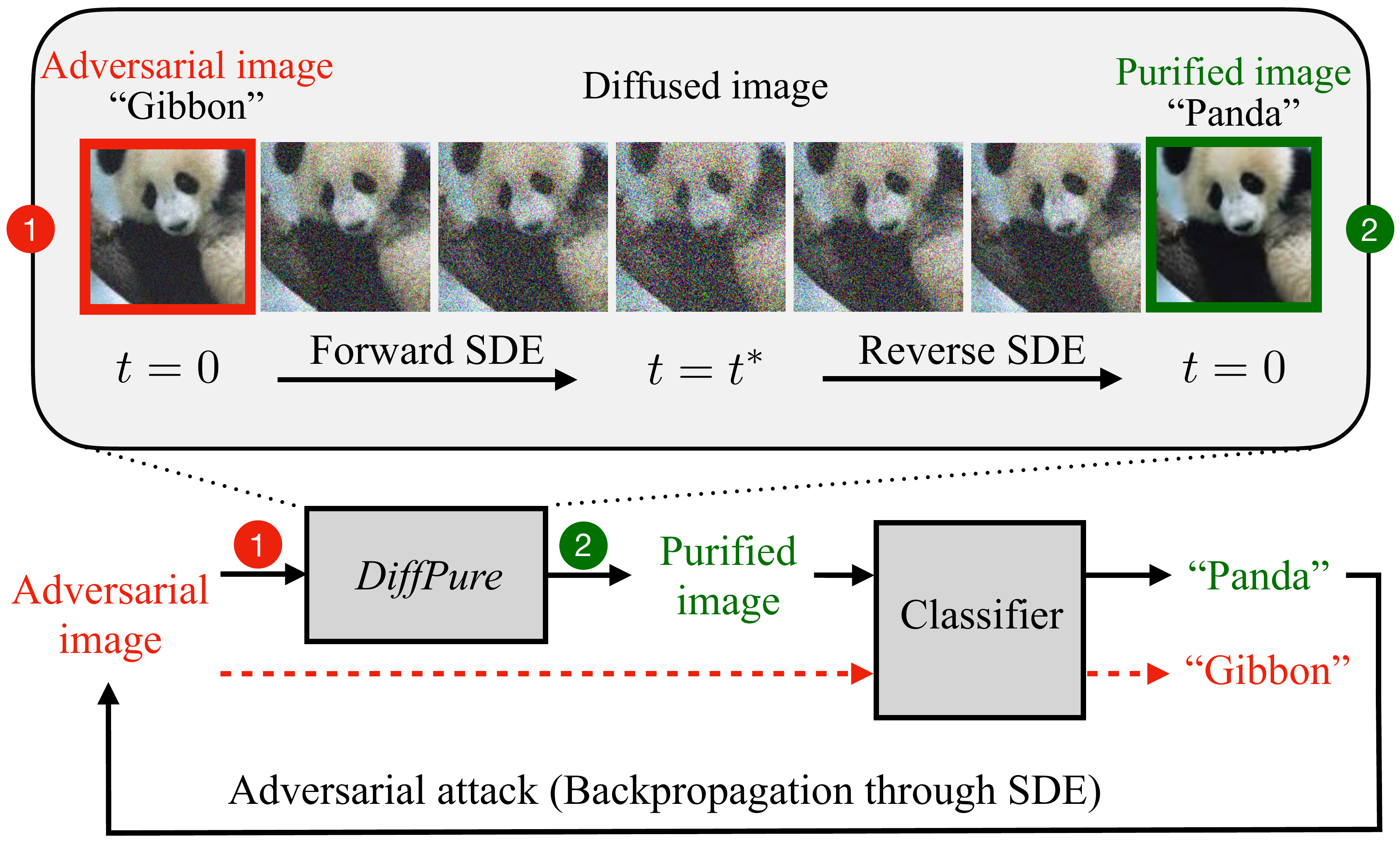}
		\vspace{-7pt}
		\caption{An illustration of \textit{DiffPure}. Given a pre-trained diffusion model, we add noise to adversarial images following the forward diffusion process with a small diffusion timestep $t^*$ to get diffused images, from which we recover clean images through the reverse denoising process before classification. Adaptive attacks backprop- agate through the SDE to get full gradients of our defense system. }
		\label{teaser}
		\vspace{-5pt}
	\end{figure}

	In this work, we propose a new adversarial purification method, termed \textit{DiffPure}, that uses the forward and reverse processes of diffusion models to purify adversarial images, as illutrated in Figure~\ref{teaser}.
	Specifically, given a pre-trained diffusion model, our method consists of two steps: (i) we first add noise to adversarial examples by following the forward process with a small diffusion timestep, and (ii) we then solve the reverse stochastic differential equation (SDE) to recover clean images from the diffused adversarial examples.
	An important design parameter in our method is the choice of diffusion timestep, since it represents the amount of noise added during the forward process. Our theoretical analysis reveals that the noise needs to be high enough to remove adversarial perturbations but not too large to destroy the label semantics of purified images. 
	Furthermore, strong adaptive attacks require gradient backpropagation through the SDE solver in our method, which suffers from the memory issue if implemented naively. Thus, we propose to use the \textit{adjoint method} to efficiently calculate full gradients of the reverse SDE with a constant memory cost.
	
	We empirically compare our method against the latest adversarial training and adversarial purification methods on various strong adaptive attack benchmarks.
	Extensive experiments on three datasets (\textit{i.e.}, CIFAR-10, ImageNet and CelebA-HQ) across multiple classifier architectures (\textit{i.e.}, ResNet, WideResNet and ViT) demonstrate the state-of-the-art performance of our method.
	For instance, compared to adversarial training methods against AutoAttack $\ell_\infty$~\citep{croce2020reliable}, our method shows absolute improvements of up to +5.44\% on CIFAR-10 and up to +7.68\% on ImageNet, respectively, in robust accuracy. 
	Moreover, compared to the latest adversarial training methods against unseen threats, our method exhibits a more significant absolute improvement (up to +36\% in robust accuracy). In comparison to adversarial purification methods against the BPDA+EOT attack~\cite{hill2021stochastic}, we have absolute improvements of +11.31\% on CIFAR-10 and +15.63\% on CelebA-HQ, respectively, in robust accuracy. 
	Finally, our ablation studies confirm the importance of noise injection in the forward and reverse processes for adversarial robustness. 
	

	\vspace{-7pt}
	\section{Background}
	\vspace{-3pt}
	
	In this section, we briefly review continuous-time diffusion models~\cite{song2021score}. 
	
	Denote by $p(\x)$ the unknown data distribution, from which each data point $\x \in \mathbb{R}^d$ is sampled. Diffusion models diffuse $p(\x)$ towards a noise distribution. The forward diffusion process $\{\x(t)\}_{t \in [0, 1]}$ is defined by an SDE with positive time increments in a fixed time horizon $[0, 1]$:
	\begin{align}\label{f_sde}
	d\x = \f(\x, t)dt + g(t)d\w,
	\end{align}
	where the initial value $\x(0):=\x \sim p(\x)$, $\f: \mathbb{R}^d \times \mathbb{R} \to \mathbb{R}^d$ is the drift coefficient, $g: \mathbb{R} \to \mathbb{R}$ is the diffusion coefficient, and $\w(t) \in \mathbb{R}^{d}$ is a standard Wiener process. 
	
	Denote by $p_t(\x)$ the marginal distribution of $\x(t)$ with $p_0(\x) := p(\x)$. In particular, $\f(\x, t)$ and $g(t)$ can be properly designed such that at the end the diffusion process, $\x(1)$ follows the standard Gaussian distribution, \textit{i.e.}, $p_1(\x) \approx \mathcal{N}(\zero, \eye_d)$.
	Throughout the paper, we consider VP-SDE~\cite{song2021score} as our diffusion model, where $\f(\x, t) := -\frac{1}{2}\beta(t)\x$ and $g(t):=\sqrt{\beta(t)}$, with $\beta(t)$ representing a time-dependent noise scale.
	By default, we use the linear noise schedule, \textit{i.e.}, $\beta(t) := \beta_{\min} + (\beta_{\max} - \beta_{\min}) t$. 
	
	Sample generation is done using the reverse-time SDE:
	\begin{align}\label{r_sde}
	d \hat{\x} = [\f(\hat{\x},t) - g(t)^2 \nabla_{\hat{\x}} \log p_t(\hat{\x})] dt + {g(t)} d\bar{\w}
	\end{align}
	where $dt$ is an infinitesimal negative time step, and $\bar{\w}(t)$ is a standard reverse-time Wiener process. Sampling $\hat{\x}(1) \sim \mathcal{N}(\zero, \eye_d)$ as the initial value and solving the above SDE from $t{=}1$ to $t{=}0$ gradually produce the less-noisy data $\hat{\x}(t)$ until we draw samples from the data distribution, \textit{i.e.}, $\hat{\x}(0) \sim p_0(\x)$. Ideally, the resulting denoising process $\{\hat{\x}(t)\}_{t \in [0, 1]}$ from Eq. (\ref{r_sde}) has the same distribution as the forward process $\{\x(t)\}_{t \in [0, 1]}$ obtained from Eq. (\ref{f_sde}). 

	The reverse-time SDE in Eq. (\ref{r_sde}) requires the knowledge of the time-dependent score function $\nabla_\x \log p_t(\x)$. One popular approach is to estimate $\nabla_\x \log p_t(\x)$ with a parameterized neural network $\s_\theta(\x, t)$~\cite{song2021score,kingma2021variational}. Accordingly, diffusion models are trained with the weighted combination of denoising score matching (DSM) across multiple time steps~\cite{vincent2011connection}: 
	\begin{align*}
	\small
	\min_\theta \! \int_0^1 \! \mathbb{E}_{p(\x)p_{0t}(\tilde{\x}|\x)} \! \left[ \lambda(t) \| \nabla_{\tilde{\x}} \log p_{0t}(\tilde{\x}|\x) \! - \! \s_\theta(\tilde{\x}, t) \|^2_2 \right] \! dt
	\end{align*}
	where $\lambda(t)$ is the weighting coefficient, 
	and $p_{0t}(\tilde{\x}|\x)$ is the transition probability from $\x(0):=\x$ to $\x(t):=\tilde{\x}$ that has a closed form through the forward SDE in Eq. (\ref{f_sde}).
	

	\vspace{-7pt}
	\section{Method}
	\label{sec:method}
	\vspace{-3pt}
	We first propose \textit{diffusion purification} (or \textit{DiffPure} for short)
	that 
	adds noise to adversarial images following the forward process of diffusion models to get diffused images, from which clean images are recovered through the reverse process. We also introduce some theoretical justifications of our method (Section~\ref{sec:diff_puri}).  Next, we apply the \textit{adjoint method} to backpropagate through SDE for efficient gradient evaluation with strong adaptive attacks (Section~\ref{sec:adapt_attack}).
	
	
	\vspace{-5pt}
	\subsection{Diffusion purification}
	\label{sec:diff_puri}
	\vspace{-3pt}
	
	Since the role of the forward SDE in Eq. (\ref{f_sde})
	is to gradually remove the local structures of data by adding noise, we hypothesize that given an adversarial example $\x_a$, if we start the forward process with $\x(0) = \x_a$, the adversarial perturbations,
	a form of small local structures added to the data, will also be gradually smoothed. 
	
	The following theorem confirms that the clean data distribution $p(\x)$ and the adversarially perturbed data distribution $q(\x)$ get closer over the forward diffusion process, implying that the adversarial perturbations will indeed be ``washed out'' by the increasingly added noise.
	
	\begin{theorem}\label{thm1}
		Let $\{\x(t)\}_{t \in [0, 1]}$ be the diffusion process defined by the forward SDE in Eq. (\ref{f_sde}). If we denote by $p_t$ and $q_t$ the respective distributions of $\x(t)$ when $\x(0) \sim p(\x)$ (\textit{i.e.}, clean data distribution) and $\x(0) \sim q(\x)$ (\textit{i.e.}, adversarial sample distribution), we then have 
		\begin{align*}
		\frac{\partial D_{KL} (p_t || q_t)}{\partial t} \leq 0
		\end{align*}
		where the equality happens only when $p_t{=}q_t$. That is, the KL divergence of $p_t$ and $q_t$ monotonically decreases when moving from $t{=}0$ to $t{=}1$ through the forward SDE. 
	\end{theorem}
	The proof follows \cite{song2021maximum,lyu2009interpretation} that build connections between Fisher divergence and the “rate of change” in KL divergence by generalizing the de Bruijn’s identity~\cite{barron1986entropy}, which we defer to Appendix~\ref{sup:proof_thm1}.
	From the above theorem, there exists a minimum timestep $t^* \in [0, 1]$ such that $D_{KL}(p_{t^*} || q_{t^*}) \leq \varepsilon $.
	However, the diffused adversarial sample $\x(t^*) \sim q_{t^*}$ at timestep $t{=}t^*$ contains additional noise and cannot be directly classified. Hence, starting from $\x(t^*)$, we can \textit{stochastically} recover the clean data at $t{=}0$ through the SDE in Eq. (\ref{r_sde}).
	
	
	\textbf{Diffusion purification: } Inspired by the observation above, we propose a two-step  adversarial purification method using diffusion models: Given an adversarial example $\x_a$ at timestep $t{=}0$,  \textit{i.e.}, $\x(0) = \x_a$, we first diffuse it by solving the forward SDE in Eq. (\ref{f_sde}) from $t{=}0$ to $t{=}t^*$.
	For VP-SDE, the diffused adversarial sample at the diffusion timestep $t^* \in [0, 1]$ can be sampled efficiently using:
	\vspace{-3pt}
	\begin{align}\label{diffused_sample}
	\x({t^*}) &= \sqrt{{\alpha}(t^*)}{\x}_a + \sqrt{1 - {\alpha}(t^*)}\eps
	\end{align}
	\vspace{-3pt}
	where $\alpha(t) =  e^{-\int_0^t \beta(s)ds}$ and $\eps \sim \mathcal{N}(\zero, \eye_d)$.
	
	Second, we solve the reverse-time SDE in Eq. (\ref{r_sde}) from the timestep $t{=}t^*$ using the diffused adversarial sample $\x(t^*)$, given by Eq. (\ref{diffused_sample}), as the initial value to get the final solution $\hat{\x}(0)$ of SDE in Eq. (\ref{r_sde}). As $\hat{\x}(0)$ does not have a closed-form solution, we resort to an SDE solver, termed \texttt{sdeint} (usually with the Euler–Maruyama discretization~\cite{kloeden1992stochastic}). That is,
	\vspace{-3pt}
	\begin{align}\label{sdeint}
	\hat{\x}(0) = \texttt{sdeint}(\x(t^*), \f_{\text{rev}}, g_{\text{rev}}, \bar{\w}, t^*, 0)
	\end{align}
	where \texttt{sdeint} is defined to sequentially take in six inputs: initial value, drift coefficient, diffusion coefficient, Wiener process, initial time, and end time. Also, the above drift and diffusion coefficients are given by
	\vspace{-3pt}
	\begin{align}\label{coeffs}
	\begin{split}
	\f_{\text{rev}}(\x, t) &:= -\frac{1}{2} \beta(t) [\x + 2 \s_\theta(\x, t)] \\
	g_{\text{rev}}(t) &:= \sqrt{\beta(t)}
	\end{split}
	\end{align}
	The resulting purified data $\hat{\x}(0)$ is then passed to an external standard classifier to make predictions. 
	An illustration of our method is shown in Figure~\ref{teaser}.
	
	\textbf{Choosing the diffusion timestep $t^*$:} 
	From Theorem~\ref{thm1},  $t^*$ should be large enough to remove local adversarial perturbations. However, $t^*$ cannot be arbitrarily large because the global label semantics will also be removed by the diffusion process if $t^*$ keeps increasing. As a result, the purified sample $\hat{\x}(0)$ cannot be classified correctly.
	
	
	Formally, the following theorem characterizes how the diffusion timestep $t^*$ affects the difference between the clean image $\x_0$ and purified image obtained by our method $\hat{\x}(0)$. 
	
	\begin{theorem}\label{thm2}
		If we assume 
		the score function satisfies that 
		$\| \s_\theta(\x,t) \| \leq \frac{1}{2} C_s$, the L2 distance between the clean data $\x$ and the purified data $\hat{\x}(0)$ given by Eq. (\ref{sdeint}) satisfies that 
		with a probability of at least $1-\delta$, we have 
		\begin{align*}
		\begin{split}
		\| \hat{\x}(0) - \x \| \leq \| \eps_a \| + \sqrt{e^{2\gamma(t^*)} - 1} C_\delta  + \gamma(t^*)C_s
		\end{split}
		\end{align*}
		where $\eps_a$ denotes the adversarial perturbation satisfying $\x_a = \x + \eps_a$, $\gamma(t^*) := \int_0^{t^*} \frac{1}{2} \beta(s)ds$ and the constant $C_\delta := \sqrt{2d + 4 \sqrt{d \log{\frac{1}{\delta}}} + 4 \log{\frac{1}{\delta}}}$. 
	\end{theorem}
	See Appendix~\ref{sup:proof_thm2} for the proof.
	Since $\gamma(t^*)$ monotonically increases with $t^*$ and $\gamma(t^*) \geq 0$ for all $t^*$, the last two terms in the above upper bound both increase with $t^*$. Thus, to make $\| \hat{\x}(0) - \x \|$ as low as possible, $t^*$ needs to be sufficiently small.
	In the extreme case where $t^*{=}0$, we have the equality that $\| \hat{\x}(0) - \x \| = \| \eps_a \|$, which means $\hat{\x}(0)$ reduces to $\x_a$ if we do not perform diffusion purification. 
	
\begin{figure}[t]
		\centering
		\begin{subfigure}{\linewidth}
			\centering
			\includegraphics[width=\linewidth]{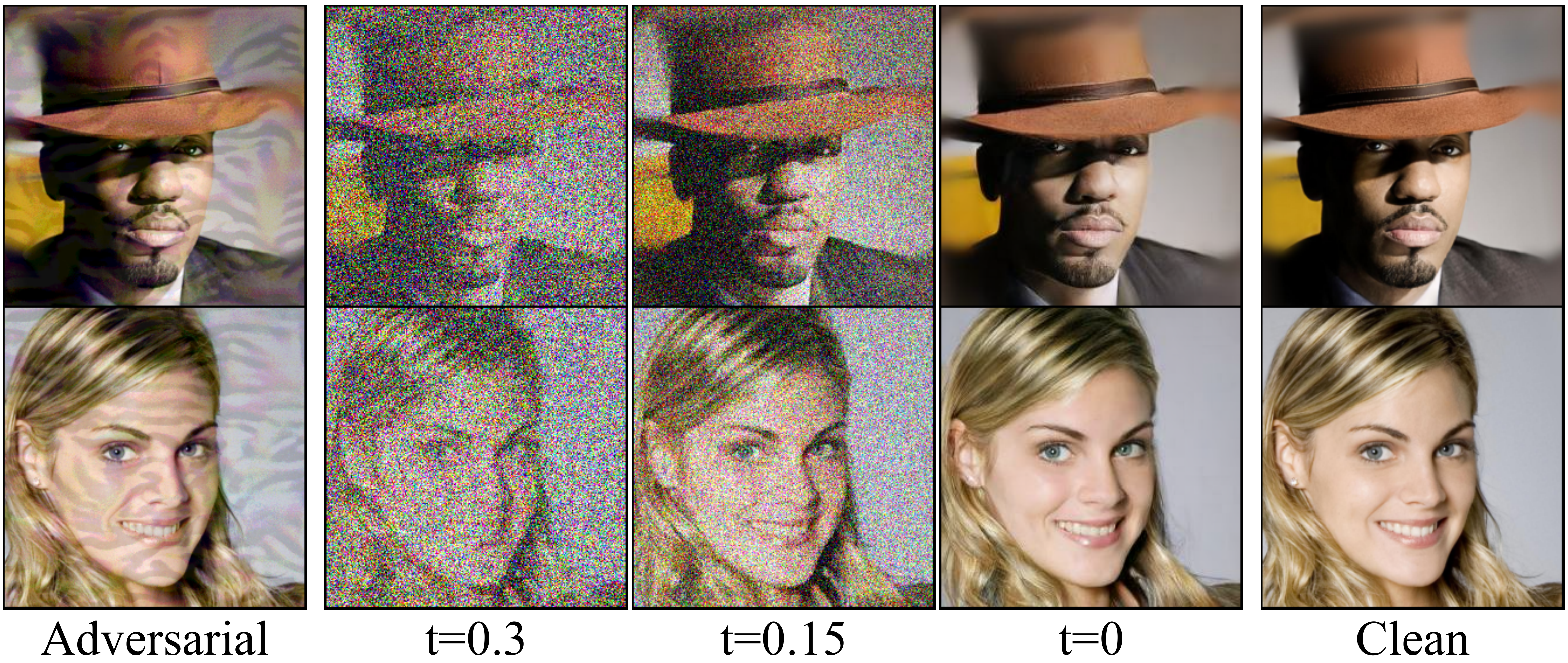}
			\vspace{-15pt}
			\caption{\small Smiling}
			\label{smile}
		\end{subfigure}
		\begin{subfigure}{\linewidth}
			\centering
			\includegraphics[width=\linewidth]{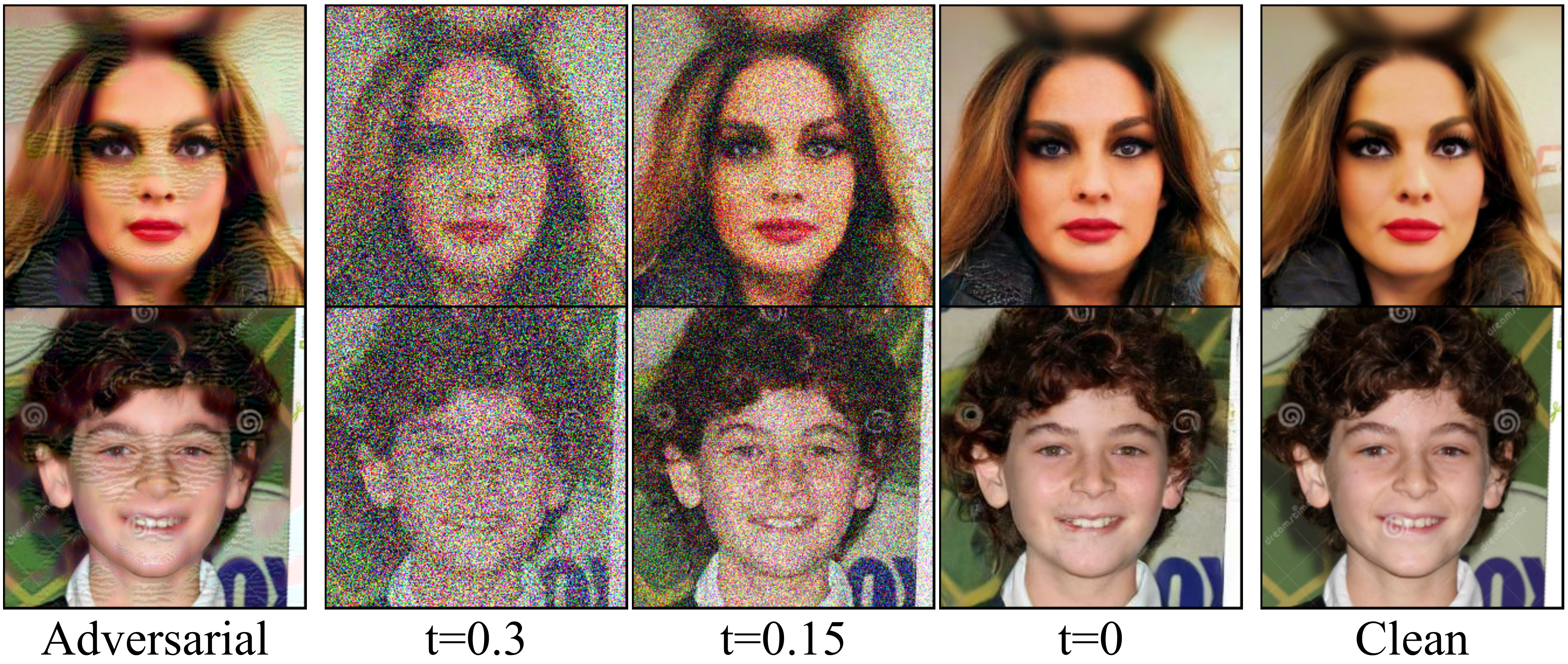}
			\vspace{-15pt}
			\caption{\small Eyeglasses}
			\vspace{-8pt}
			\label{eyeglasses}
		\end{subfigure}
		\caption{Our method purifies adversarial examples (first column) produced by attacking attribute classifiers using PGD $\ell_\infty$ ($\epsilon=16/255$), where $t^*=0.3$. The middle three columns show the results of the SDE in Eq. (\ref{sdeint}) at different timesteps, and we observe the purified images at $t{=}0$ match the clean images (last column). Better zoom in to see how we remove adversarial perturbations.
		}
		\label{vis_adv}
		\vspace{-7pt}
	\end{figure}

	Due to the trade-off between purifying the local perturbations (with a larger $t^*$) and preserving the global structures (with a smaller $t^*$) of adversarial examples, there exists a sweet spot for the diffusion timestep $t^*$ to obtain a high robust classification accuracy. 
	Since adversarial perturbations are usually small, which can be removed with a small $t^*$, the best $t^*$ in most adversarial robustness tasks also remain relatively small.
	As a proof of concept, we provide visual examples in Figure \ref{vis_adv} to show how our method purifies the adversarial perturbations while maintaining the global semantic structures.
	See Appendix \ref{sup:sec_purify_vis} for more results. 
	
	

	\vspace{-4pt}
	\subsection{Adaptive attack to diffusion purification}
	\label{sec:adapt_attack}
	\vspace{-2pt}
	
	
	Strong adaptive attacks~\cite{athalye2018obfuscated,tramer2020adaptive} require computing full gradients of our defense system. However, simply backpropagating through the SDE solver in Eq. (\ref{sdeint}) scales poorly in the computational memory. In particular, denote by $N$ the number of function evaluations in solving the SDE, the required memory increases by $\mathcal{O}(N)$. This issue makes it challenging to effectively evaluate our method with strong adaptive attacks. 
	
	Prior adversarial purification methods~\citep{shi2020online,yoon2021adversarial} suffer from the same memory issue with strong adaptive
	attacks. Thus, they either evaluate only with black-box  attacks or change the evaluation strategy to circumvent the full gradient computation (\textit{e.g.}, using approximate gradients). This makes them difficult to compare with adversarial training methods under the more standard evaluation protocols (\textit{e.g.}, AutoAttack). To overcome this, we propose to use the \textit{adjoint method}~\cite{li2020scalable} to efficiently compute full gradients of the SDE without the memory issue. The intuition is that the gradient through an SDE can be obtained by solving another augmented SDE. 
	
	The following proposition provides the augmented SDE for calculating the gradient of an objective $\mathcal{L}$ w.r.t. the input ${x}(t^*)$ of the SDE in Eq. (\ref{sdeint}).
	\begin{proposition}\label{prop1}
		For the SDE in Eq. (\ref{sdeint}), the augmented SDE that computes the gradient $\frac{\partial \mathcal{L}}{\partial {\x}(t^*)}$ of backpropagating through it is given by
		\begin{align}\label{aug_sdeint}
		\renewcommand\arraystretch{1.5}
		\! \begin{pmatrix}
		{\x}(t^*) \\
		\frac{\partial \mathcal{L}}{\partial {\x}(t^*)} 
		\end{pmatrix}
		\! = \! \texttt{sdeint} \! \left( 
		\! \begin{pmatrix}
		\hat{\x}(0) \\ 
		\frac{\partial \mathcal{L}}{\partial \hat{\x}(0)} 
		\end{pmatrix}\!,
		\tilde{\f}, \tilde{\g}, \tilde{\w}, 0, t^*
		\right)
		\end{align}
		where $\frac{\partial \mathcal{L}}{\partial \hat{\x}(0)}$ is the gradient of the objective $\mathcal{L}$ w.r.t. the output $\hat{\x}(0)$ of the SDE in Eq. (\ref{sdeint}), and
		\begin{align*}
		\begin{split}
		\tilde{\f}([\x; \z], t) &=
		\renewcommand\arraystretch{1.5}
		\begin{pmatrix}
		\f_{\text{rev}}(\x, t) \\
		{\frac{\partial \f_{\text{rev}}(\x, t) }{\partial \x}} \z
		\end{pmatrix}  \\
		\renewcommand\arraystretch{1.5}
		\tilde{\g}(t) &=
		\begin{pmatrix}
		-g_{\text{rev}}(t) \one_d \\
		\zero_d
		\end{pmatrix} \\
		\renewcommand\arraystretch{1.5}
		\tilde{\w}(t) &=
		\begin{pmatrix}
		-\w(1-t) \\
		-\w(1-t)
		\end{pmatrix}
		\end{split}
		\end{align*}
		with $\one_d$ and $\zero_d$ representing the $d$-dimensional vectors of all ones and all zeros, respectively.
	\end{proposition}
	The proof is deferred to Appendix~\ref{sec:aug_sdeint_supp}.
	Ideally if the SDE solver has a small numerical error, the gradient obtained from this proposition will closely match its true value (see Appendix~\ref{grad_ana}).
	As the gradient computation has been converted to solving the augmented SDE in Eq. (\ref{aug_sdeint}), we do not need to store intermediate operations and thus end up with the $\mathcal{O}(1)$ memory cost~\cite{li2020scalable}.
	That is, the adjoint method described above turns the reverse-time SDE 
	in Eq. (\ref{sdeint}) into a differentiable operation (without the memory issue). Since the forward diffusion step in Eq. (\ref{diffused_sample}) is also differentiable using the reparameterization trick, we can easily compute full gradients of a loss function regarding the adversarial images for strong adaptive attacks.

	\vspace{-8pt}
	\section{Related work}
	\label{related_work}
	\vspace{-2pt}
	
	
	\vspace{-2pt}
	\textbf{Adversarial training} It learns a robust classifier by training on adversarial examples created during every weight update.
	After first introduced by \citet{madry2018towards}, adversarial training has become one of the most successful defense methods in neural networks against adversarial attacks~\citep{gowal2020uncovering,rebuffi2021fixing}. 
	Despite the difference in the defense form, some variants of adversarial training share similarities with our method. \citet{he2019parametric} inject Gaussian noise to each network layer for better robustness via stochastic effects. \citet{kang2021stable} train neural ODEs with Lyapunov-stable equilibrium points for adversarial defense. \citet{gowal2021improving} use generative models for data augmentation to improve adversarial training, where diffusion models work the best.
	
	
	\textbf{Adversarial purification}
	Using generative models to purify adversarial images before classification, adversarial purification has become a promising counterpart of adversarial training. \citet{samangouei2018defense} propose defense-GAN using GANs as the purification model, and \citet{song2018pixeldefend} propose PixelDefense by relying on autoregressive generative models. More recently, \citet{du2019implicit,grathwohl2020your,hill2021stochastic} show the improved robustness of using EBMs to purify attacked images via Langevin dynamics (LD). More similarly, \citet{yoon2021adversarial} use the denoising score-based model~\cite{song2019generative} for purification, but its sampling is still a variant of LD that does not rely on forward diffusion and backward denoising processes.
	We empirically compare our method against these previous works and we largely outperform them.

	\vspace{-12pt}
	\paragraph{Diffusion models}
	As a probabilistic generative models for unsupervised modeling~\cite{ho2020denoising}, diffusion models have shown strong sample quality and diversity in image synthesis~\citep{dhariwal2021diffusion,song2021maximum}. Since then, they have been used in many image editing tasks, such as image-to-image translation~\citep{meng2021sdedit,choi2021ilvr,saharia2021palette} and text-guided image editing~\cite{kim2021diffusionclip,nichol2021glide}.
	Although adversarial purification can be considered as a special image editing task and particularly DiffPure shares a similar procedure with SDEdit~\citep{meng2021sdedit}, none of these works apply diffusion models to improve the model robustness. Besides, evaluating our method with strong adaptive attacks poses a new challenge of backpropagating through the denoising process that previous works do not deal with.

	\vspace{-8pt}
	\section{Experiments}
	\label{exp}
	\vspace{-3pt}
	
	In this section, we first provide experimental settings (Section~\ref{exp_set}). On various strong adaptive attack benchmarks, we then compare our method with the state-of-the-art adversarial training and adversarial purification methods (Section~\ref{comp_adv_train} to \ref{comp_purify}). We defer the results against standard attack (\textit{i.e.}, non-adaptive) and black-box attack, suggested by \citet{croce2022evaluating}, to Appendix \ref{app_sec_stand_black} for completeness.
	Next, we perform various ablation studies to provide better insights into our method (Section~\ref{ablation}).
	
	
	\vspace{-6pt}
	\subsection{Experimental settings}
	\label{exp_set}
	\vspace{-3pt}
	
	\paragraph{Datasets and network architectures}
	We consider three datasets for evaluation: CIFAR-10~\cite{krizhevsky2009learning}, CelebA-HQ~\cite{karras2018progressive}, and ImageNet~\cite{deng2009imagenet}. Particularly, we compare with the state-of-the-art defense methods reported by the standardized benchmark RobustBench~\cite{croce2020robustbench} on CIFAR-10 and ImageNet while comparing with other adversarial purification methods on CIFAR-10 and CelebA-HQ following their settings. 
	For classifiers, we consider three widely used architectures: ResNet~\cite{he2016deep}, WideResNet~\cite{zagoruyko2016wide} and ViT~\cite{dosovitskiy2021image}.
	
	
	\vspace{-7pt}
	\paragraph{Adversarial attacks} 
	We evaluate our method with strong adaptive attacks. We use the commonly used AutoAttack $\ell_\infty$ and $\ell_2$ threat models~\cite{croce2020reliable} to compare with adversarial training methods. To show the broader applicability of our method beyond $\ell_p$-norm attacks, we also evaluate with the spatially
	transformed adversarial examples (StAdv)~\cite{xiao2018spatially}.
	Due to the stochasticity introduced by the diffusion and denoising processes (Section \ref{sec:diff_puri}),
	we apply Expectation Over Time (EOT)~\cite{athalye2018obfuscated} to these adaptive attacks, where we use EOT=20 (see Figure~\ref{eot} for more details). Besides, we apply the BPDA+EOT attack~\cite{hill2021stochastic} to make a fair comparison with other adversarial purification methods.
	
	\vspace{-7pt}
	\paragraph{Evaluation metrics}
	We consider two metrics to evaluate the performance of defense approaches: \textit{standard accuracy} and \textit{robust accuracy}. The standard accuracy measures the performance of the defense method on clean data, which is evaluated on the whole test set in each dataset. The robust accuracy measures the performance on adversarial examples generated by adaptive attacks. 
	Due to the high computational cost of applying adaptive attacks to our method, unless stated otherwise, we evaluate robust accuracy for our method and previous works on a fixed subset of 512 images randomly sampled from the test set. Notably, robust accuracies of most baselines do not change much on the sampled subset, compared to the whole test set (see Appendix~\ref{rob_acc_from_ours_vs_}). 
	
	We defer more details of  the above experimental settings and the baselines that we compare with to Appendix~\ref{sec:exp_set_supp}.
	
	
	\begin{table}[t] 
		\vspace{-5pt}
		\centering
		\caption{Standard accuracy and robust accuracy against AutoAttack $\ell_\infty$ ($\epsilon={8}/{255}$) on CIFAR-10, obtained by different classifier architectures. In our method, the diffusion timestep is $t^*=0.1$. }
		\label{tab_L_inf_cifar10}
		\footnotesize\addtolength{\tabcolsep}{-4pt}
		\vspace{3pt}
		\begin{tabular}{cccc}
			\hline
			Method & Extra Data & Standard Acc & Robust Acc \\
			\hline & \\[-2ex]
			\rowcolor{LightCyan}
			WideResNet-28-10 & & &
			\\
			\hline
			\text{ \cite{zhang2020geometry}} & \cmark & 89.36 & 59.96
			\\
			\text{ \cite{wu2020adversarial}} & \cmark & 88.25 & 62.11
			\\
			\text{ \cite{gowal2020uncovering}} & \cmark & 89.48 & 62.70
			\\
			\hline
			\text{ \cite{wu2020adversarial}} & \xmark & 85.36 & 59.18
			\\
			\text{ \cite{rebuffi2021fixing}} & \xmark & 87.33 & 61.72
			\\
			\text{ \cite{gowal2021improving}} & \xmark & 87.50 & 65.24 \\ 
			Ours & \xmark & \textbf{89.02\text{$\pm$0.21}} & \textbf{70.64\text{$\pm$0.39}}
			\\
			\hline & \\[-2ex]
			\rowcolor{LightCyan}
			WideResNet-70-16 & & &
			\\
			\hline
			\text{ \cite{gowal2020uncovering}} & \cmark & 91.10 & 66.02
			\\
			\text{ \cite{rebuffi2021fixing}} & \cmark & 92.23 & 68.56
			\\
			\hline
			\text{ \cite{gowal2020uncovering}} & \xmark & 85.29 & 59.57
			\\
			\text{ \cite{rebuffi2021fixing}} & \xmark & 88.54 & 64.46
			\\ 
			\text{ \cite{gowal2021improving}} & \xmark & 88.74 & 66.60
			\\ 
			Ours & \xmark & \textbf{90.07\text{$\pm$0.97}} & \textbf{71.29\text{$\pm$0.55}}
			\\
			\hline
		\end{tabular}
		\vspace{-8pt}
	\end{table}

	\begin{table}[t] 
		\vspace{-5pt}
		\centering
		\caption{Standard accuracy and robust accuracy against AutoAttack $\ell_2$ ($\epsilon=0.5$) on CIFAR-10, obtained by different classifier architectures. In our method, the diffusion timestep is $t^*=0.075$. ($^*$Methods use WideResNet-34-10, with the same width but more layers than the default one.)}
		\label{tab_L2_cifar10}
		\footnotesize\addtolength{\tabcolsep}{-4pt}
		\vspace{3pt}
		\begin{tabular}{cccc}
			\hline
			Method & Extra Data & Standard Acc & Robust Acc \\
			\hline & \\[-2ex]
			\rowcolor{LightCyan}
			WideResNet-28-10 & & &
			\\ \hline
			\text{ \cite{augustin2020adversarial}}$^*$ & \cmark & 92.23 & 77.93
			\\ \hline
			\text{ \cite{rony2019decoupling}} & \xmark & 89.05 & 66.41 \\ 
			\text{ \cite{ding2020mma}} & \xmark & 88.02 & 67.77 \\ 
			\text{ \cite{wu2020adversarial}}$^*$ & \xmark & 88.51 & 72.85
			\\
			\text{ \cite{sehwag2021robust}}$^*$ & \xmark & 90.31 & 75.39
			\\
			\text{ \cite{rebuffi2021fixing}} & \xmark & \textbf{91.79} & {78.32} \\ 
			Ours & \xmark & {91.03\text{$\pm$0.35}} & \textbf{78.58\text{$\pm$0.40}}
			\\
			\hline & \\[-2ex]
			\rowcolor{LightCyan}
			WideResNet-70-16 & & & \\ \hline
			\text{ \cite{gowal2020uncovering}} & \cmark & 94.74 & 79.88
			\\
			\text{ \cite{rebuffi2021fixing}} & \cmark & 95.74 & 81.44
			\\ 
			\hline
			\text{ \cite{gowal2020uncovering}} & \xmark & 90.90 & 74.03
			\\
			\text{ \cite{rebuffi2021fixing}} & \xmark & {92.41} & \textbf{80.86} \\ 
			Ours & \xmark & \textbf{92.68\text{$\pm$0.56}} & {80.60\text{$\pm$0.57}}
			\\
			\hline
		\end{tabular}
		\vspace{-13pt}
	\end{table}

	\begin{table}[t] 
		\vspace{-5pt}
		\centering
		\caption{Standard accuracy and robust accuracy against AutoAttack $\ell_\infty$  ($\epsilon={4}/{255}$) on ImageNet, obtained by different classifier architectures. In our method, the diffusion timestep is $t^*=0.15$. 
			($^\dagger$Robust accuracy is directly reported from the respective paper.)
		}
		\label{tab_imagenet}
		\footnotesize\addtolength{\tabcolsep}{-4pt}
		\vspace{3pt}
		\begin{tabular}{cccc}
			\hline
			Method & Extra Data & Standard Acc & Robust Acc \\
			\hline & \\[-2ex]
			\rowcolor{LightCyan}
			ResNet-50 & & & \\ \hline
			\text{ \cite{robustness}} & \xmark & 62.56 & 31.06
			\\
			\text{ \cite{wong2020fast}} & \xmark & 55.62 & 26.95
			\\
			\text{ \cite{salman2020adversarially}} & \xmark & 64.02 & 37.89
			\\
			\text{ \cite{bai2021transformers}}$^\dagger$ & \xmark & 67.38 & 35.51
			\\
			Ours & \xmark & \textbf{67.79\text{$\pm$0.43}} & \textbf{40.93\text{$\pm$1.96}}
			\\
			\hline & \\[-2ex]
			\rowcolor{LightCyan}
			WideResNet-50-2 & & & \\ \hline
			\text{ \cite{salman2020adversarially}} & \xmark & 68.46 & 39.25
			\\
			Ours & \xmark & \textbf{71.16\text{$\pm$0.75}} & \textbf{44.39\text{$\pm$0.95}}
			\\
			\hline & \\[-2ex]
			\rowcolor{LightCyan}
			DeiT-S & & & \\ \hline
			\text{ \cite{bai2021transformers}}$^\dagger$ & \xmark & 66.50 & 35.50 
			\\ 
			Ours & \xmark & \textbf{73.63\text{$\pm$0.62}} & \textbf{43.18\text{$\pm$1.27}}
			\\
			\hline
		\end{tabular}
		\vspace{-10pt}
	\end{table}


	\begin{table*}[t] 
		\vspace{-5pt}
		\centering
		\caption{Standard accuracy and robust accuracies against unseen threat models on ResNet-50 for CIFAR-10. We keep the same evaluation settings with \cite{laidlaw2021perceptual}, where the attack bounds are $\epsilon=8/255$ for AutoAttack $\ell_\infty$, $\epsilon=1$ for AutoAttack $\ell_2$, and $\epsilon=0.05$ for StAdv. The baseline results are reported from the respective papers. For our method, the diffusion timestep is $t^*=0.125$.}
		\label{unseen_threat}
		\footnotesize\addtolength{\tabcolsep}{0pt}
		\vspace{3pt}
		
		\begin{tabular}{ccccc}
			\hline
			
			\multirow{2}{*}{Method} &
			\multirow{2}{*}{Standard Acc} &
			\multicolumn{3}{c}{Robust Acc}
			\\
			\cline{3-5} 
			& & $\ell_\infty$ & $\ell_2$ & StAdv \\
			
			\hline
			Adv. Training with $\ell_\infty$~\cite{laidlaw2021perceptual} & 86.8 & \textcolor{gray}{49.0} & 19.2 & 4.8
			\\
			Adv. Training with $\ell_2$~\cite{laidlaw2021perceptual} & 85.0 & 39.5 & \textcolor{gray}{47.8} & 7.8 
			\\
			Adv. Training with StAdv~\cite{laidlaw2021perceptual} & 86.2 & 0.1 & 0.2 & \textcolor{gray}{53.9}
			\\
			\hline
			PAT-self~\cite{laidlaw2021perceptual} & 82.4 & 30.2 & 34.9 & 46.4
			\\
			\textsc{Adv. Craig}~\citep{dolatabadi2021ell_} & 83.2 & 40.0 & 33.9 & 49.6 \\
			\textsc{Adv. GradMatch}~\citep{dolatabadi2021ell_} & 83.1 & 39.2 & 34.1 & 48.9 \\
			\hline
			Ours & \textbf{88.2\text{$\pm$0.8}} & \textbf{70.0$\pm$1.2} &
			\textbf{70.9$\pm$0.6} &
			\textbf{55.0\text{$\pm$0.7}}
			\\
			\hline
		\end{tabular}
		\vspace{-7pt}
	\end{table*}
	
	\vspace{-4pt}
	\subsection{Comparison with the state-of-the-art}
	\label{comp_adv_train}
	\vspace{-3pt}
	
	We first compare DiffPure with the state-of-the-art adversarial training methods reported by RobustBench~\citep{croce2020robustbench}, 
	against the $\ell_\infty$ and $\ell_2$ threat models, respectively. 
	
	\vspace{-6pt}
	\paragraph{CIFAR-10}
	Table \ref{tab_L_inf_cifar10} shows the robustness performance against $\ell_\infty$ threat model ($\epsilon=8/255$) with AutoAttack on CIFAR-10. We can see that our method achieves both better standard accuracy and better robust accuracy than previous state-of-the-art methods that do not use extra data on different classifier architectures. In specific, our method improves robust accuracy by 5.44\% on WideResNet-28-10 and by 4.69\% on WideResNet-70-16, respectively.
	Furthermore, our method even largely outperforms baselines trained \textit{with extra data} regarding robust accuracies, with comparable standard accuracies with different classifiers. 
	
	Table \ref{tab_L2_cifar10} shows the robustness performance against $\ell_2$ threat model ($\epsilon=0.5$) with AutoAttack on CIFAR-10. We can see that our method outperforms most defense methods without using extra data while being on par with the best performing method~\cite{rebuffi2021fixing}, regarding both standard and robust accuracies. 
	The gap between our method and \cite{rebuffi2021fixing} trained with extra data exists, but can be leveled up by replacing the standard classifier in our method with the adversarially trained one, as shown in Figure~\ref{comb_at}.
	
	These results demonstrate the effectiveness of our method in defending against $\ell_\infty$ and $\ell_2$ threat models on CIFAR-10. 
	It is worth noting that in contrast to the competing methods that are trained for the specific $\ell_p$-norm attack used in evaluation, our method is agnostic to the threat model.

	
	\vspace{-8pt}
	\paragraph{ImageNet}
	
	Table \ref{tab_imagenet} shows the robustness performance against $\ell_\infty$ threat model ($\epsilon=4/255$) with AutoAttack on ImageNet.
	We evaluate our method on two CNN architectures: ResNet-50 and WideResNet-50-2, and one ViT architecture: DeiT-S~\cite{pmlr-v139-touvron21a}. 
	We can see that our method largely outperforms the state-of-the-art baselines regarding both the standard and robust accuracies. Besides, the advantages of our method over baselines become more significant on the ViT architecture. Specifically, our method improves robust accuracy by 3.04\% and 5.14\% on ResNet-50 and WideResNet-50-2, respectively, and by 7.68\% on DeiT-S. For standard accuracy on DeiT-S, our method also largely improves over the baseline by 7.13\%.
	
	These results clearly demonstrate the effectiveness of our method in defending against $\ell_\infty$ threat models on ImageNet.
	Note that for the adversarial training baselines, the training recipes for CNNs cannot be directly applied to ViTs due to the over-regularization issue~\citep{bai2021transformers}. However, our method is agnostic to classifier architectures.
	


	\begin{table*}[t] 
		\vspace{-5pt}
		\centering
		\caption{Comparison with other adversarial purification methods using the BPDA+EOT attack with $\ell_\infty$ perturbations. (a) We evaluate on the \textit{eyeglasses} attribute classifier for CelebA-HQ, where $\epsilon=16/255$. See Appendix~\ref{sec:comp_celeba_supp} for similar results on the \textit{smiling} attribute. Note that \textsc{Opt} and \textsc{Enc} denote the optimization-based and econder-based GAN inversions, respectively, and \textsc{Enc}+\textsc{Opt} implies a combination of \textsc{Opt} and \textsc{Enc}.
			(b) We evaluate on WideResNet-28-10 for CIFAR-10, and keep the experimental settings the same with \cite{hill2021stochastic}, where $\epsilon=8/255$. 
			($^*$The purification is actually a variant of the LD sampling.)
		}
		\begin{subtable}[b]{0.52\linewidth}
			\footnotesize\addtolength{\tabcolsep}{-4.4pt}
			\centering
			\vspace{-5pt}
			\caption{CelebA-HQ}
			\label{tab_celeba_purify}
			\vspace{-5pt}
			\begin{tabular}{ccccc}
				
				\hline
				Method & Purification & Standard Acc & Robust Acc \\
				\hline
				\cite{vahdat2020NVAE} & VAE & \textbf{99.43} & 0.00 \\
				\cite{karras2020analyzing} & GAN+\textsc{Opt} & 97.76 & 10.80 \\
				\cite{chai2021ensembling} & GAN+\textsc{Enc}+\textsc{Opt} & {99.37} & 26.37 \\
				\cite{richardson2021encoding} & GAN+\textsc{Enc} & 93.95 & 75.00 \\
				\hline
				Ours ($t^* = 0.4$) & Diffusion & 93.87$\pm$0.18 & 89.47$\pm$1.18 \\
				Ours ($t^* = 0.5$) & Diffusion & 93.77$\pm$0.30 & \textbf{90.63$\pm$1.10}
				\\
				\hline
			\end{tabular}
		\end{subtable}
		\quad
		\begin{subtable}[b]{0.45\linewidth}
			\footnotesize\addtolength{\tabcolsep}{-4.4pt}
			\centering
			\vspace{5pt}
			\caption{CIFAR-10}
			\label{tab_cifar10_purify}
			\vspace{-5pt}
			\begin{tabular}{ccccc}
				\hline
				Method & Purification & Standard Acc & Robust Acc \\
				\hline
				\cite{song2018pixeldefend} & Gibbs Update & \textbf{95.00} & 9.00 \\
				\cite{yang2019me} & Mask+Recon. & 94.00 & 15.00 \\
				\cite{hill2021stochastic} & EBM+LD & 84.12 & 54.90 \\
				\cite{yoon2021adversarial} & DSM+LD$^*$ & 86.14 & 70.01 \\
				\hline
				Ours ($t^* = 0.075$) & Diffusion & 91.03$\pm$0.35 & 77.43$\pm$0.19 \\
				Ours ($t^* = 0.1$) & Diffusion & 89.02$\pm$0.21 & \textbf{81.40$\pm$0.16}
				\\
				\hline
			\end{tabular}
		\end{subtable}
		\vspace{-7pt}
	\end{table*}

	\vspace{-5pt}
	\subsection{Defense against unseen threats}
	\label{unseen_defense}
	\vspace{-3pt}
	
	The main drawback of the adversarial training baselines is their poor generalization to unseen attacks: even if models are robust against a specific threat model, they are still fragile against other threat models.
	To see this, we evaluate each method with three attacks: $\ell_\infty$, $\ell_2$ and StAdv, shown in Table~\ref{unseen_threat}. 
	Note that for the plain adversarial training methods with a specific attack objective (\textit{e.g.}, Adv Train - $\ell_\infty$), only other threat models (\textit{e.g.}, $\ell_2$ and StAdv) are considered unseen. We thus mark the seen threats by gray.
	
	We can see that our method is robust to all three unseen threat models while the performances of these plain adversarial baselines drop significantly against unseen attacks.
	Compared with the state-of-the-art defense methods against unseen threat models~\citep{laidlaw2021perceptual,dolatabadi2021ell_}, our method achieves significantly better standard accuracy and robust accuracies across all three attacks. In particular, the robust accuracy of our method improves by 30\%, 36\% and 5.4\% on $\ell_\infty$, $\ell_2$ and StAdv, respectively.

	\vspace{-5pt}
	\subsection{Comparison with other purification methods}
	\label{comp_purify}
	\vspace{-3pt}
	
	Because most prior adversarial purification methods have an optimization or sampling loop in their defense process~\cite{hill2021stochastic}, they cannot be evaluated directly with the strongest white-box adaptive attacks, such as AutoAttack. 
	To this end, we use the BPDA+EOT attack~\cite{tramer2020adaptive,hill2021stochastic}, an adaptive attack designed specifically for purification methods (with stochasticity), to evaluate our method and baselines for a fair comparison. 
	
	\vspace{-9pt}
	\paragraph{CelebA-HQ}
	We compare with other strong generative models, such as NVAE~\cite{vahdat2020NVAE} and StyleGAN2~\cite{karras2020analyzing}, that can be used to purify adversarial examples. The basic idea is to first encode adversarial images to latent codes, with which purified images are synthesized from the decoder (see Appendix~\ref{sec:basline_supp} for implementation details). We choose CelebA-HQ for the comparsion because they both perform well on it. 
	In Table~\ref{tab_celeba_purify}, we use the \textit{eyeglasses} attribute to show that our method has much better robust accuracy (+15.63\%) than the best performing baseline while also maintaining a relatively high standard accuracy. We defer the similar results on the \textit{smiling} attribute to Appendix~\ref{sec:comp_celeba_supp}. These results demonstrate the superior performance of diffusion models in adversarial robustness than other generative models as a purification model.
	
	
	\vspace{-8pt}
	\paragraph{CIFAR-10}
	In Table \ref{tab_cifar10_purify}, we compare our method with other adversarial purification methods on CIFAR-10, where the methods based on the LD sampling for purification are the state-of-the-art~\cite{hill2021stochastic,yoon2021adversarial}. We observe that our method largely outperforms previous methods against the BPDA+EOT attack, with an absolute improvement of at least +11.31\% in robust accuracy. Meanwhile, we can slightly trade-off robust accuracy for better standard accuracy by decreasing $t^*$, making it comparable to the best reported standard accuracy (\textit{i.e.}, 91.03\% vs. 95.00\%). 
	These results show that our method becomes a new state-of-the-art in adversarial purification.
	
	
	\vspace{-5pt}
	\subsection{Ablation studies}
	\label{ablation}
	\vspace{-3pt}
	
	\begin{figure}[t]
		\centering
		\includegraphics[width=0.72\linewidth]{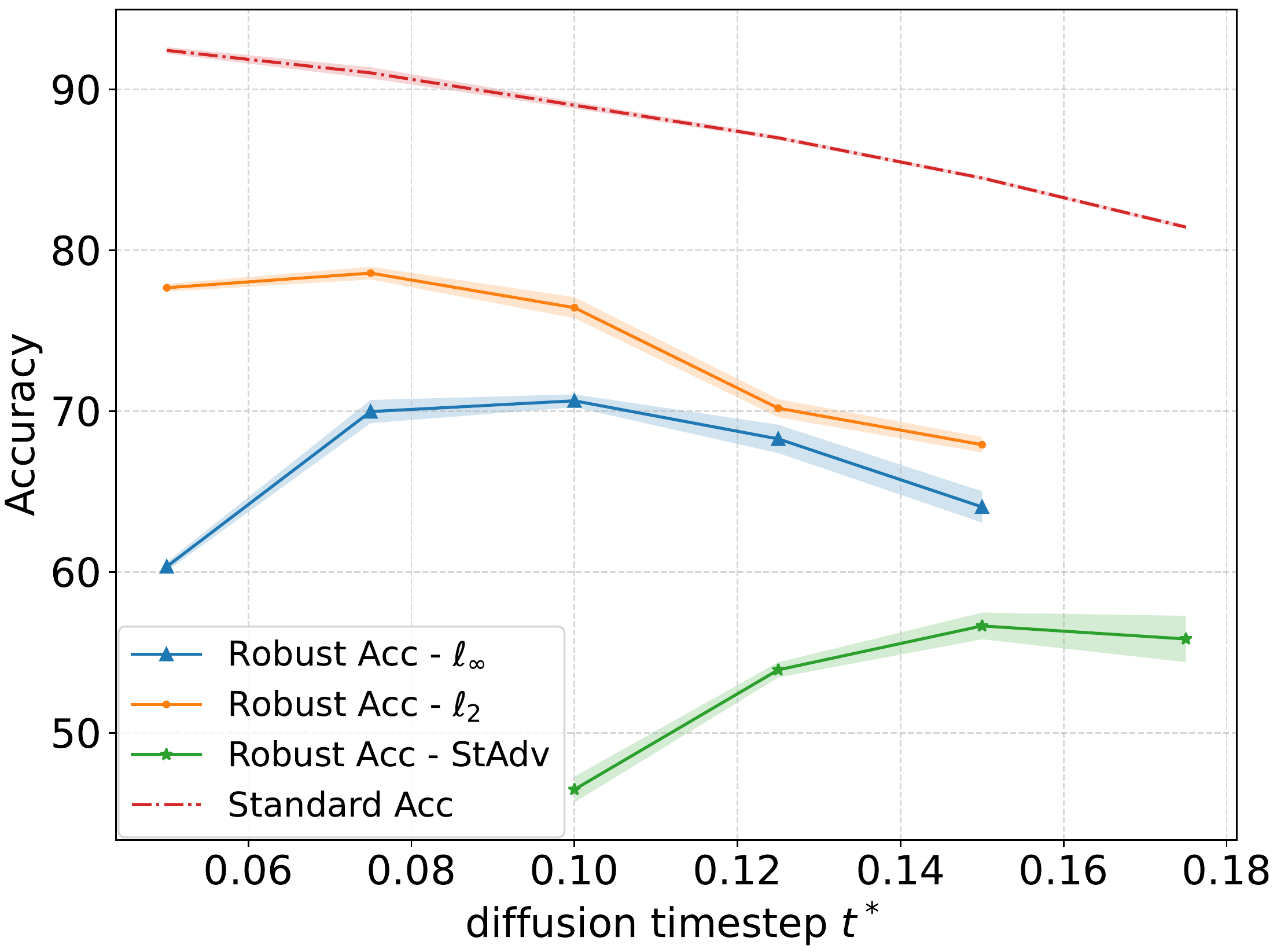}
		\vspace{-8pt}
		\caption{\small Impact of diffusion time $t^*$ in our method on standard accuracy and robust accuracies against AutoAttack $\ell_\infty$ ($\epsilon=8/255$), $\ell_2$ ($\epsilon=0.5$) and StAdv ($\epsilon=0.05$) threat models, respectively, where we evaluate on WideResNet-28-10 for CIFAR-10.}
		\label{impact_t}
		\vspace{-6pt}
	\end{figure}
	
	\paragraph{Impact of diffusion timestep $t^*$}
	We first show how the diffusion timestep $t^*$ affects the robustness performance of our method against different threat models in Figure \ref{impact_t}. 
	We can see that (i) the standard accuracy monotonically decreases with $t^*$ since more label semantics are lost with the larger diffusion timestep, and (ii) all the robust accuracies first increase and then decrease as $t^*$ becomes larger due to the trade-off as discussed in Section~\ref{sec:diff_puri}. 
	Notably, the optimal timestep $t^*$ for the best robust accuracy remains small but also varies across different threat models (\textit{e.g.},  $\ell_\infty$: $t^*{=}0.075$, $\ell_2$: $t^*{=}0.1$, and StAdv: $t^*{=}0.15$). 
	Since stronger perturbations need a larger diffusion timestep to be smoothed, it implies that StAdv ($\epsilon{=}0.05$) perturbs the input images the most while $\ell_2$ ($\epsilon{=}0.5$) does the least. 
	
	\vspace{-10pt}
	\paragraph{Impact of sampling strategy}
	Given the pre-trained diffusion models, besides relying on VP-SDE, there are other ways of recovering clean images from the adversarial examples. Here we consider another two sampling strategies: (i) LD-SDE (\textit{i.e.}, an SDE formulation of the LD sampling that samples from an EBM, formed by our score function at $t{=}0$), and (ii) VP-ODE (\textit{i.e.}, an equivalent ODE sampling derived from VP-SDE that solves the reverse generative process using the probability flow ODEs \cite{song2021score}).
	Please see Appendix \ref{sec:other_sampling} for more details about these sampling variants. In Table \ref{tab_sampl_strategy}, we compare different sampling strategies with the same diffusion model. 
	
	Although each sampling strategy has a comparable standard accuracy, our method achieves a significantly better robust accuracy.
	To explain this, we hypothesize that (i) the LD sampling only uses the score function with clean images at timestep $t{=}0$, making it less robust to noisy (or perturbed) input images, while our method considers score functions at various noise levels. (ii) The ODE sampling introduces much less randomness to the defense model, due to its deterministic trajectories, and thus is more vulnerable to adaptive attacks from the randomized smoothing perspective~\citep{cohen2019certified, pinot2020randomization}. 
	Inspired by this, we can add more stochasticity by using a randomized diffusion timestep $t^*$ for the improved performance (see Appendix~\ref{sec:abalation_supp}).
	
	\begin{table}[t] 
		\vspace{-8pt}
		\centering
		\caption{We compare different sampling strategies by evaluating on WideResNet-28-10 with AutoAttack $\ell_\infty$ ($\epsilon=8/255$) for CIFAR-10. We use $t^*=0.1$ for both VP-ODE and VP-SDE (Ours), while using the best hyperparameters after grid search for LD-SDE.}
		\label{tab_sampl_strategy}
		\footnotesize\addtolength{\tabcolsep}{0pt}
		\vspace{3pt}
		\begin{tabular}{cccc}
			\hline
			Sampling & Standard Acc & Robust Acc \\
			\hline
			LD-SDE & 87.36$\pm$0.09 & 38.54$\pm$1.55 \\
			VP-ODE & \textbf{90.79$\pm$0.12} & 39.86$\pm$0.98 \\
			\hline
			VP-SDE (Ours) & 89.02$\pm$0.21 & \textbf{70.64$\pm$0.39}
			\\
			\hline
		\end{tabular}
		\vspace{-8pt}
	\end{table}

	
	\begin{figure}[t]
		\centering
		\includegraphics[width=0.68\linewidth]{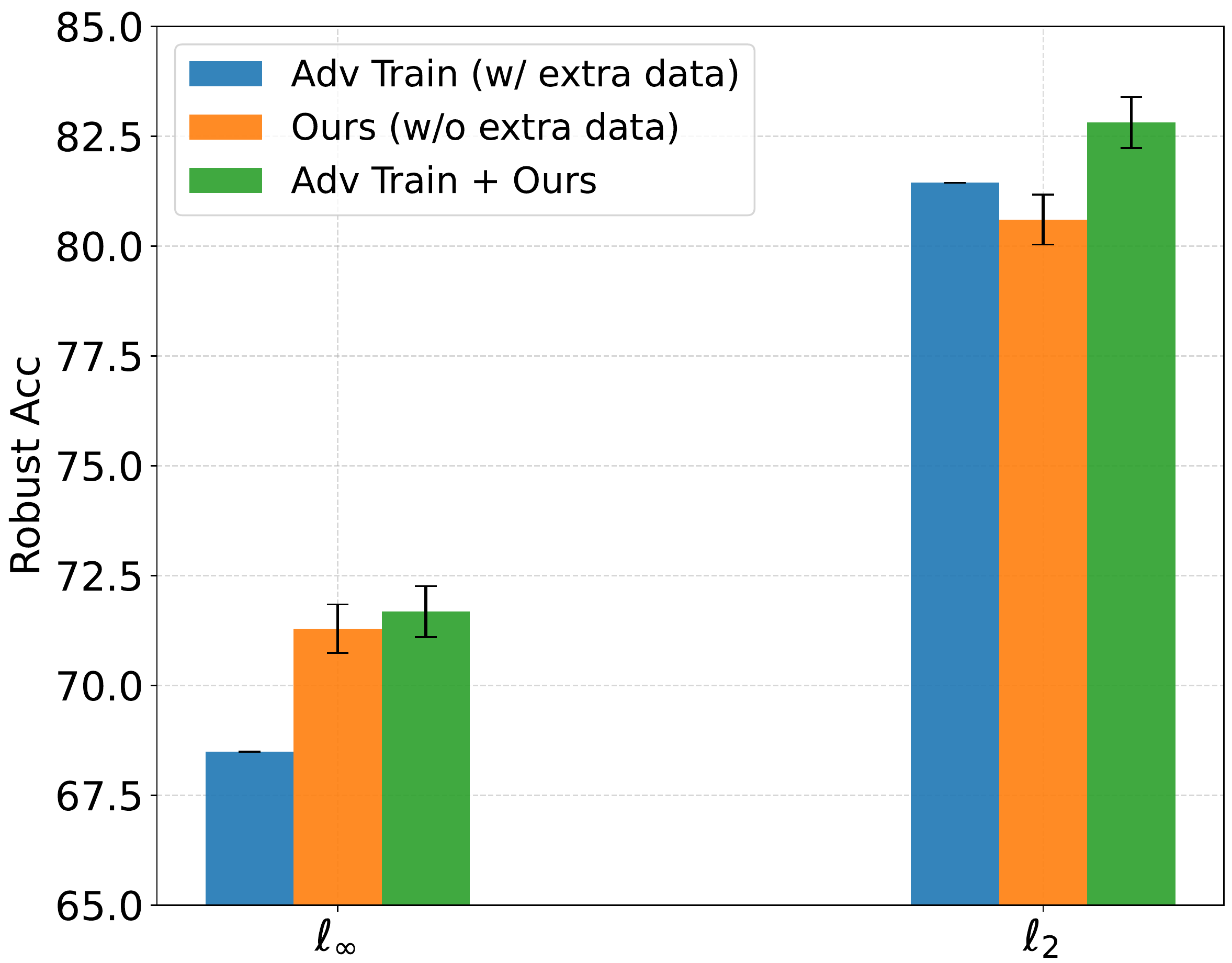}
		\vspace{-8pt}
		\caption{\small Combination of our method with adversarial training, where we evaluate on WideResNet-76-10 for CIFAR-10 with AutoAttack $\ell_\infty$ ($\epsilon=8/255$) and $\ell_2$ ($\epsilon=0.5$) threat models, respectively. Regarding adversarial training, we use the model in \cite{rebuffi2021fixing} that is adversarially trained with extra data.}
		\vspace{-12pt}
		\label{comb_at}
	\end{figure}
	
	\vspace{-10pt}
	\paragraph{Combination with adversarial training}
	Since our proposed \textit{DiffPure} is an orthogonal defense method to adversarial training, we can also combine our method with adversarial training (\textit{i.e.}, feeding the purified images from our method to the adversarially trained classifiers).
	Figure \ref{comb_at} shows that this combination (\textit{i.e.}, ``Adv Train + Ours'') can improve the robust accuracies against AutoAttack $\ell_\infty$ and $\ell_2$ threat models, respectively.
	Besides, by comparing the results against the $\ell_\infty$ and $\ell_2$ threat models, the improvement from the combination over our method with the standard classifier (\textit{i.e.}, ``Ours'') becomes more significant when the adversarial training method with extra data (\textit{i.e.}, ``Adv Train'') is already on par with our method.
	Therefore, 
	we can apply our method to the pre-existing adversarially trained classifiers for further improving the performance. 

	
	
	
	\vspace{-5pt}
	\section{Conclusions}
	\label{concl}
	\vspace{-3pt}
	
	We proposed a new defense method called \textit{DiffPure} that applies diffusion models to purify adversarial examples before feeding them into classifiers. We also applied the adjoint method to compute full gradients of the SDE solver for evaluating with strong white-box adaptive attacks. To show the robustness performance of our method, we conducted extensive experiments on CIFAR-10, ImageNet and CelebA-HQ with different classifiers architectures including ResNet, WideResNet and ViT to compare with the state-of-the-art adversarial training and adversarial purification methods. In defense of various strong adaptive attacks such as AutoAttack, StAdv and BPDA+EOT, our method largely outperforms previous approaches. 
	
	Despite the large improvements, our method has two major limitations: (i) the purification process takes much time (proportional to the diffusion timestep, see Appendix~\ref{app_infer_time}), making our method inapplicable to the real-time tasks, and (ii) diffusion models are sensitive to image colors, making our method incapable of defending color-related corruptions. It is interesting to either apply recent works on accelerating diffusion models or design new diffusion models specifically for model robustness to overcome these two limitations.

	
	\bibliography{diffusion_defense}
	\bibliographystyle{icml2022}

	\newpage
	\appendix
	\onecolumn
	
	\section{Proofs in Section~\ref{sec:method}}
	
	\subsection{Proof of Theorem~\ref{thm1}}
	\label{sup:proof_thm1}
	
	\begin{theorem}\label{thm1_supp}
		Let $\{\x(t)\}_{t \in [0, 1]}$ be the diffusion process defined by the forward SDE in Eq. (\ref{f_sde}). If we denote by $p_t$ and $q_t$ the respective distributions of $\x(t)$ when $\x(0) \sim p(\x)$ (\textit{i.e.}, clean data distribution) and $\x(0) \sim q(\x)$ (\textit{i.e.}, adversarial sample distribution), we then have 
		\begin{align}
		\frac{\partial D_{KL} (p_t || q_t)}{\partial t} \leq 0
		\end{align}
		where the equality happens only when $p_t{=}q_t$. That is, the KL divergence of $p_t$ and $q_t$ monotonically decreases when moving from $t{=}0$ to $t{=}1$ through the forward SDE. 
	\end{theorem}
	
	\emph{Proof:} The proof follows \cite{song2021maximum}.
	First, the Fokker-Planck equation \cite{sarkka2019applied} for the forward SDE in Eq. (\ref{f_sde}) is given by 
	\begin{align}\label{fpe}
	\begin{split}
	\frac{\partial p_t(\x)}{\partial t} &= -\nabla_\x \cdot \left( f(\x, t)p_t(\x) - \frac{1}{2}g^2(t) \nabla_\x p_t(\x)  \right) \\
	&= -\nabla_\x \cdot \left( f(\x, t) p_t(\x) - \frac{1}{2}g^2(t) p_t(\x) \nabla_\x \log p_t(\x)  \right) \\
	&= \nabla_\x \cdot \left( \h_p(\x,t) p_t(\x) \right)
	\end{split}
	\end{align}
	where we define $\h_p(\x, t) := \frac{1}{2}g^2(t) \nabla_\x \log p_t(\x) -f(\x, t) $.
	Then if we assume $p_t(x)$ and $q_t(x)$ are smooth and fast decaying, \textit{i.e.},
	\begin{align}\label{smooth_decay_assump}
	\lim_{\x_i \to \infty} p_t(\x) \frac{\partial }{\partial \x_i} \log p_t(\x) = 0 \quad \text{and} \quad \lim_{\x_i \to \infty} q_t(\x) \frac{\partial }{\partial \x_i} \log q_t(\x) = 0,
	\end{align}
	for any $i = 1,\cdots, d$, we can evaluate
	\begin{align*}
	\begin{split}
	\frac{\partial D_{KL} (p_t || q_t)}{\partial t} &= \frac{\partial}{\partial t} \int p_t(\x) \log \frac{p_t(\x)}{q_t(\x)} d\x \\
	&= \int \frac{\partial p_t(\x)}{\partial t} \log \frac{p_t(\x)}{q_t(\x)} d\x + \underbrace{ \int \frac{\partial p_t(\x)}{\partial t} d\x }_{=0} + \int \frac{p_t(\x)}{q_t(\x)} \frac{\partial q_t(\x)}{\partial t} d\x \\
	& \mathop{=}^{(a)} \int \nabla_\x \cdot \left( \h_p(\x,t) p_t(\x) \right) \log \frac{p_t(\x)}{q_t(\x)} d\x + \int \frac{p_t(\x)}{q_t(\x)} \nabla_\x \cdot \left( \h_q(\x,t) q_t(\x) \right) d\x \\
	&\mathop{=}^{(b)} -\int p_t(\x) [\h_p(\x,t)  - \h_q(\x,t)]^T[\nabla_\x \log p_t(\x) - \nabla_\x \log q_t(\x)] d\x \\
	&= -\frac{1}{2}g^2(t)  \int p_t(\x) \| \nabla_\x \log p_t(\x) - \nabla_\x \log q_t(\x) \|^2_2 d\x \\
	&\mathop{=}^{(c)} -\frac{1}{2}g^2(t) D_F(p_t || q_t)
	\end{split}
	\end{align*}
	where $(a)$ follows by plugging Eq. (\ref{fpe}), $(b)$ follows from the integration by parts and the assumption in Eq. (\ref{smooth_decay_assump}), and $(c)$ follows from the definition of the \textit{Fisher divergence}: $D_F(p_t || q_t) := \int p_t(\x) \| \nabla_\x \log p_t(\x) - \nabla_\x \log q_t(\x) \|^2_2 d\x$. 
	
	Since $g^2(t) > 0$, and the Fisher divergence satisfies that $D_F(p_t || q_t) \geq 0$ and $D_F(p_t || q_t) = 0$ if and only if $p_t = q_t$, we have  
	\begin{align*}
	\frac{\partial D_{KL} (p_t || q_t)}{\partial t} \leq 0
	\end{align*}
	where the equality happens only when $p_t{=}q_t$. \hfill $\square$
	
	\vspace{-4pt}
	\subsection{Proof of Theorem~\ref{thm2}}
	\label{sup:proof_thm2}
	\vspace{-2pt}
	
	\begin{theorem}\label{thm2_supp}
		If we assume 
		the score function satisfies that 
		$\| \s_\theta(\x,t) \| \leq \frac{1}{2} C_s$, the L2 distance between the clean data $\x$ and the purified data $\hat{\x}(0)$ given by Eq. (\ref{sdeint}) satisfies that if 
		with a probability of at least $1-\delta$, we have 
		\begin{align}
		\begin{split}
		\| \hat{\x}(0) - \x \| \leq \| \eps_a \| + \sqrt{e^{2\gamma(t^*)} - 1} C_\delta  + \gamma(t^*)C_s
		\end{split}
		\end{align}
		where $\gamma(t^*) := \int_0^{t^*} \frac{1}{2} \beta(s)ds$ and the constant $C_\delta := \sqrt{2d + 4 \sqrt{d \log{\frac{1}{\delta}}} + 4 \log{\frac{1}{\delta}}}$.
	\end{theorem}
	
	\emph{Proof:} 
	Denote by $\eps_a$ the adversarial perturbation, we have the adversarial example ${\x}_a = \x + \eps_a$, where $\x$ represents the clean image. 
	Because the diffused adversarial example $x(t^*)$ through the forward diffusion process satisfies 
	\begin{align}\label{diffused_sample_supp}
	\x({t^*}) = \sqrt{{\alpha}(t^*)}{\x}_a + \sqrt{1 - {\alpha}(t^*)}\eps_1
	\end{align}
	where $\alpha(t) =  e^{-\int_0^t \beta(s)ds}$ and $\eps_1 \sim \mathcal{N}(\zero, \eye_d)$, the L2 distance between the clean data $x_0$ and the purified data $\hat{x}(0)$ can be bounded as
	\begin{align}
	\begin{split}
	\| \hat{\x}(0) - \x \| &= \| \x(t^*) + (\hat{\x}(0) - \x(t^*)) - \x \| \\
	&= \| \x(t^*) +\int_{t^*}^0 -\frac{1}{2} \beta(t) [\x + 2 \s_\theta(\x, t)] dt + \int_{t^*}^0\sqrt{\beta(t)} d\bar{\w} - \x \| \\
	& \leq \| \underbrace{\x(t^*) +\int_{t^*}^0 -\frac{1}{2} \beta(t) \x dt + \int_{t^*}^0\sqrt{\beta(t)} d\bar{\w}}
	_{\text{Integration of Linear SDE}} - \; \x \| + \|\int_{t^*}^0-\beta(t) \s_\theta(\x, t) dt \| \\
	\end{split}
	\end{align}
	where the second equation follows from the integration of the reverse-time SDE defined in Eq. (\ref{sdeint}), and in the last line we have separated the integration of the linear SDE from non-linear SDE involving the score function $\s_\theta(\x, t)$ by using the triangle inequality.

	The above linear SDE is a time-varying Ornstein–Uhlenbeck process with a negative time increment that starts from $t{=}t^*$ to $t{=}0$ with the initial value set to $\x(t^*)$. 
	Denote by $\x'(0)$ its solution, from \cite{sarkka2019applied} we know $\x'(0)$ follows a Gaussian distribution, where its mean $\bmu(0)$ and covariance matrix $\bSigma(0)$ are the solutions of the following two differential equations, respectively:
	\begin{align}
	\frac{d \bmu}{dt} &= -\frac{1}{2} \beta(t) \bmu \\
	\frac{d \bSigma}{dt} &= -\beta(t) \bSigma + \beta(t) \eye_d
	\end{align}
	with the initial conditions $\bmu(t^*)=\x(t^*)$ and $\bSigma(t^*)=\zero$.
	By solving these two differential equations, we have that conditioned on $\x(t^*)$, $\x'(0) \sim \mathcal{N}(e^{\gamma(t^*)}\x(t^*), (e^{2\gamma(t^*)} - 1)\eye_d)$, where $\gamma(t^*) := \int_0^{t^*} \frac{1}{2} \beta(s)ds$. 
	
	Using the reparameterization trick, we have:
	\begin{equation}
	\begin{split}
	\x'(0) - \x &= e^{\gamma(t^*)}\x(t^*) + \sqrt{e^{2\gamma(t^*)} - 1} \eps_2 - \x \\
	&=e^{\gamma(t^*)}\left(e^{-\gamma(t^*)}(\x + \eps_a) + \sqrt{1 - e^{-2 \gamma(t)}} \eps_1\right)  + \sqrt{e^{2\gamma(t^*)} - 1} \eps_2 - \x \\
	&= \sqrt{e^{2\gamma(t^*)} - 1} (\eps_1 + \eps_2) + \eps_a
	\end{split}
	\end{equation}
	where the the second equation follows by substituting Eq. (\ref{diffused_sample_supp}).
	Since $\eps_1$ and $\eps_2$ are independent, the first term can be represented as a single zero-mean Normal variable with the variance $2(e^{2\gamma(t^*)} - 1)$.
	Assuming that the norm of the score function $\s_\theta(\x,t)$ is bounded by a constant $\frac{1}{2}C_s$ and $\eps \sim \mathcal{N}(\zero, \eye_d)$, we have:
	\begin{align}
	\begin{split}
	\| \hat{\x}(0) - \x \| \leq \| \sqrt{2(e^{2\gamma(t^*)} - 1)} \eps  + \eps_a \| + \gamma(t^*)C_s
	\end{split}
	\end{align}
	Since $\|\eps \|^2 \sim \chi^2(d)$, from the concentration inequality~\cite{boucheron2013concentration}, we have 
	\begin{align}
	\text{Pr}(\|\eps \|^2 \geq d + 2 \sqrt{d \sigma} + 2 \sigma) \leq e^{-\sigma}
	\end{align}
	Let $e^{-\sigma} = \delta$, we get 
	\begin{align}
	\text{Pr} \left( \|\eps \| \geq \sqrt{d + 2 \sqrt{d \log{\frac{1}{\delta}}} + 2 \log{\frac{1}{\delta}}} \right) \leq \delta
	\end{align}
	Therefore, with the probability of at least $1-\delta$, we have 
	\begin{align}
	\begin{split}
	\| \hat{\x}(0) - \x \| \leq \| \eps_a \| + \sqrt{e^{2\gamma(t^*)} - 1} C_\delta  + \gamma(t^*)C_s
	\end{split}
	\end{align}
	where the constant $C_\delta: = \sqrt{2d + 4 \sqrt{d \log{\frac{1}{\delta}}} + 4 \log{\frac{1}{\delta}}}$.
	\hfill $\square$

	\vspace{-8pt}
	\subsection{Proof of Proposition~\ref{prop1}}
	\label{sec:aug_sdeint_supp}
	\vspace{-2pt}
	
	\begin{proposition}\label{prop1_supp}
		For the SDE in Eq. (\ref{sdeint}), the augmented SDE that computes the gradient $\frac{\partial \mathcal{L}}{\partial {\x}(t^*)}$ of backpropagating through it is given by
		\begin{align}\label{aug_sdeint_supp}
		\renewcommand\arraystretch{1.5}
		\! \begin{pmatrix}
		{\x}(t^*) \\
		\frac{\partial \mathcal{L}}{\partial {\x}(t^*)} 
		\end{pmatrix}
		\! = \! \texttt{sdeint} \! \left( 
		\! \begin{pmatrix}
		\hat{\x}(0) \\ 
		\frac{\partial \mathcal{L}}{\partial \hat{\x}(0)} 
		\end{pmatrix}\!,
		\tilde{\f}, \tilde{\g}, \tilde{\w}, 0, t^*
		\right)
		\end{align}
		where $\frac{\partial \mathcal{L}}{\partial \hat{\x}(0)}$ is the gradient of the objective $\mathcal{L}$ w.r.t. the output $\hat{\x}(0)$ of the SDE in Eq. (\ref{sdeint}), and
		\begin{align*}
		\begin{split}
		\tilde{\f}([\x; \z], t) &=
		\renewcommand\arraystretch{1.5}
		\begin{pmatrix}
		\f_{\text{rev}}(\x, t) \\
		{\frac{\partial \f_{\text{rev}}(\x, t) }{\partial \x}} \z
		\end{pmatrix}  \\
		\renewcommand\arraystretch{1.5}
		\tilde{\g}(t) &=
		\begin{pmatrix}
		-g_{\text{rev}}(t) \one_d \\
		\zero_d
		\end{pmatrix} \\
		\renewcommand\arraystretch{1.5}
		\tilde{\w}(t) &=
		\begin{pmatrix}
		-\w(1-t) \\
		-\w(1-t)
		\end{pmatrix}
		\end{split}
		\end{align*}
		with $\one_d$ and $\zero_d$ representing the $d$-dimensional vectors of all ones and all zeros, respectively.
	\end{proposition}

	\emph{Proof:}
	Before applying the adjoint method, we first transform the reverse-time SDE in Eq. (\ref{sdeint}) to a forward SDE, by a change of variable $t':=1 - t$ such that $t' \in [1-t^*, 1] \subset [0,1]$.
	With this, the equivalent forward SDE with positive time increments from $t{=}1-t^*$ to $t{=}1$ becomes
	\begin{align}\label{fwd_sdeint_sup}
	\hat{\x}(0) = \texttt{sdeint}(\x({t^*}), {\f}_{\text{fwd}}, {g}_{\text{fwd}}, \w, 1-t^*, 1)
	\end{align}
	where the drift and diffusion coefficients are
	\begin{align*}
	\begin{split}
	{\f}_{\text{fwd}}(\x, t) &= -\f_{\text{rev}}(\x, 1-t) \\
	{g}_{\text{fwd}}(t) &= g_{\text{rev}}(1-t)
	\end{split}
	\end{align*}
	By following the stochastic adjoint method proposed in \cite{li2020scalable}, the augmented SDE that computes the gradient $\frac{\partial \mathcal{L}}{\partial {\x}(t^*)}$ of the objective $\mathcal{L}$ w.r.t. the input $\x(t^*)$ of the SDE in Eq. (\ref{fwd_sdeint_sup}) is given by
	\begin{align}\label{aug_sdeint_supp_v1}
	\renewcommand\arraystretch{1.5}
	\! \begin{pmatrix}
	{\x}(t^*) \\
	\frac{\partial \mathcal{L}}{\partial {\x}(t^*)} 
	\end{pmatrix}
	\! = \! \texttt{sdeint} \! \left( 
	\! \begin{pmatrix}
	\hat{\x}(0) \\ 
	\frac{\partial \mathcal{L}}{\partial \hat{\x}(0)} 
	\end{pmatrix}\!,
	\tilde{\f}_{\text{fwd}}, \tilde{\g}_{\text{fwd}}, \tilde{\w}_{\text{fwd}}, -1, t^*\!-\!1
	\right)
	\end{align}
	where $\frac{\partial \mathcal{L}}{\partial \hat{\x}(0)}$ is the gradient of the objective $\mathcal{L}$ w.r.t. the output $\hat{\x}(0)$ of the SDE in Eq. (\ref{fwd_sdeint_sup}), and the augmented drift coefficient $\tilde{\f}_{\text{fwd}}: \mathbb{R}^{2d} \times \mathbb{R} \to \mathbb{R}^{2d}$, the augmented diffusion coefficient $\tilde{\g}_{\text{fwd}}: \mathbb{R} \to \mathbb{R}^{2d}$ and the augmented Wiener process $\tilde{\w}(t) \in \mathbb{R}^{2d}$ are given by
	\begin{align*}
	\begin{split}
	\tilde{\f}_{\text{fwd}}([\x; \z], t) &=
	\renewcommand\arraystretch{1.5}
	\begin{pmatrix}
	-\f_{\text{fwd}}(\x, -t) \\
	{\frac{\partial \f_{\text{fwd}}(\x, -t) }{\partial \x}} \z
	\end{pmatrix} = \renewcommand\arraystretch{1.5}
	\begin{pmatrix}
	\f_{\text{rev}}(\x, 1+t) \\
	{\frac{\partial \f_{\text{rev}}(\x, 1+t) }{\partial \x}} \z
	\end{pmatrix} \\
	\renewcommand\arraystretch{1.5}
	\tilde{\g}_{\text{fwd}}(t) &= 
	\begin{pmatrix}
	-g_{\text{fwd}}(-t) \one_d \\
	\zero_d
	\end{pmatrix} = \begin{pmatrix}
	-g_{\text{rev}}(1+t) \one_d \\
	\zero_d
	\end{pmatrix} \\
	\renewcommand\arraystretch{1.5}
	\tilde{\w}_{\text{fwd}}(t) &= 
	\begin{pmatrix}
	-\w(-t) \\
	-\w(-t)
	\end{pmatrix}
	\end{split}
	\end{align*}
	with $\one_d$ and $\zero_d$ representing the $d$-dimensional vectors of all ones and all zeros, respectively.
	Note that the augmented SDE in Eq. (\ref{aug_sdeint_supp_v1}) moves from $t=-1$ to $t=t^*-1$. Similarly, with the change of variable $t':= 1 + t$ such that $t' \in [0, t^*]$, we can rewrite the augmented SDE as 
	\begin{align}\label{aug_sdeint_supp_v2}
	\renewcommand\arraystretch{1.5}
	\! \begin{pmatrix}
	{\x}(t^*) \\
	\frac{\partial \mathcal{L}}{\partial {\x}(t^*)} 
	\end{pmatrix}
	\! = \! \texttt{sdeint} \! \left( 
	\! \begin{pmatrix}
	\hat{\x}(0) \\ 
	\frac{\partial \mathcal{L}}{\partial \hat{\x}(0)} 
	\end{pmatrix}\!,
	\tilde{\f}, \tilde{\g}, \tilde{\w}, 0, t^*
	\right)
	\end{align}
	where
	\begin{align*}
	\begin{split}
	\tilde{\f}([\x; \z], t) &= \tilde{\f}_{\text{fwd}}([\x; \z], t-1) =
	\renewcommand\arraystretch{1.5}
	\begin{pmatrix}
	\f_{\text{rev}}(\x, t) \\
	{\frac{\partial \f_{\text{rev}}(\x, t) }{\partial \x}} \z
	\end{pmatrix} \\
	\renewcommand\arraystretch{1.5}
	\tilde{\g}(t) &= \tilde{\g}_{\text{fwd}}(t-1) =
	\begin{pmatrix}
	-g_{\text{rev}}(t) \one_d \\
	\zero_d
	\end{pmatrix} \\
	\renewcommand\arraystretch{1.5}
	\tilde{\w}(t) &= \tilde{\w}_{\text{fwd}}(t-1) =
	\begin{pmatrix}
	-\w(1-t) \\
	-\w(1-t)
	\end{pmatrix}
	\end{split}
	\end{align*}
	and $\f_{\text{rev}}$ and $g_{\text{rev}}$ are given by Eq. (\ref{coeffs}). \hfill $\square$

	\section{More details of experimental settings}
	\label{sec:exp_set_supp}

	\subsection{Implementation details of our method}
	
	First, our method requires solving two SDEs: a reverse-time denosing SDE in Eq. (\ref{sdeint}) to get purified images, and an augmented SDE in Eq. (\ref{aug_sdeint}) to compute gradients through the SDE in Eq. (\ref{sdeint}). In experiments, we use the adjoint framework for SDEs named \texttt{adjoint\_sdeint} in the TorchSDE library: \url{https://github.com/google-research/torchsde} for both adversarial purification and gradient evaluation. We use the simple Euler-Maruyama method to solve both SDEs with a fixed step size \texttt{dt}=$10^{-3}$. 
	Ideally, the step size should be as small as possible to ensure that our gradient computation has an infinitely small numerical error. 
	However, small steps sizes come with a high computational cost due to the increase in the number of neural network evaluations. 
	We empirically observe that the robust accuracy of our method barely change any more if we further reduce the step size from $10^{-3}$ to $10^{-4}$ during a sanity check. Hence, we use a step size of $10^{-3}$ for all experiments to save the time in the purification process through the SDE solver. Note that this step size is often used in the denoising diffusion models as well \cite{ho2020denoising,song2021score}.
	
	Second, our method also requires the pre-trained diffusion models. In experiments, we use different pre-trained models on three datasets: Score SDE~\cite{song2021score} for CIFAR-10, Guided Diffusion~\cite{dhariwal2021diffusion} for ImageNet and DDPM~\cite{ho2020denoising} for CelebA-HQ. In specific, we use the \texttt{vp/cifar10\_ddpmpp\_deep\_continuous} checkpoint from the \texttt{score\_sde} library: \url{https://github.com/yang-song/score_sde} for the CIFAR-10 experiments. We use the \texttt{256x256 diffusion (unconditional)} checkpoint from the \texttt{guided-diffusion} library: \url{https://github.com/openai/guided-diffusion} for the ImageNet experiments. Finally, for the CelebA-HQ experiments, we use the CelebA-HQ checkpoint from the \texttt{SDEdit} library: \url{https://github.com/ermongroup/SDEdit}.

	\subsection{Implementation details of adversarial attacks}
	
	\paragraph{AutoAttack}
	We use AutoAttack to compare with the state-of-the-art adversarial training methods, as reported in the RobustBench benchmark. To make a fair comparison, we uses their codebase: \url{https://github.com/RobustBench/robustbench}  with default hyperparameters for evaluation. Similarly, we set $\epsilon=8/255$ and $\epsilon=0.5$ for AutoAttack $\ell_\infty$ and AutoAttack $\ell_2$, respectively, on CIFAR-10. For AutoAttack $\ell_\infty$ on ImageNet, we set $\epsilon=4/255$. 
	
	There are two versions of AutoAttack: (i) the \textsc{Standard} version, which contains four attacks: APGD-CE, APGD-T, FAB-T and Square, and is mainly used for evaluating deterministic defense methods, and (ii) the \textsc{Rand} version, which contains two attacks: APGD-CE and APGD-DLR, and is used for evaluating stochastic defense methods.
	Because there is stochasticity in our method, we consider the \textsc{Rand} version and choose the default EOT=20 for both $\ell_\infty$ and $\ell_2$ after searching for the minimum EOT with which the robust accuracy does not further decrease (see Figure~\ref{eot}).
	
	In practice, we find that in a few cases, the \textsc{Standard} version actually makes a stronger attack (indicated by a lower robust accuracy) to our method than the \textsc{Rand} version. Therefore, to measure the worse-case defense performance of our method, we run both the \textsc{Standard} version and the \textsc{Rand} version of AutoAttack, and report the minimum robust accuracy of these two versions as our final robust accuracy.
	
	\paragraph{StAdv}
	We use the StAdv attack to demonstrate that our method can defend against unseen threats beyond $\ell_p$-norm attacks. We closely follow the codebase of PAT~\cite{laidlaw2021perceptual}: \url{https://github.com/cassidylaidlaw/perceptual-advex} with default hyperparameters for evaluation. Moreover, we add the EOT to average out the stochasticity of gradients in our defense method. Similarly, we use EOT=20 by default for the StAdv attack after searching for the minimum EOT that the robust accuracy saturates (see Figure~\ref{eot}).

	\paragraph{BPDA+EOT}
	For many adversarial purification methods where there exists an optimization loop or non-differentiable operations, the BPDA attack is known as the strongest attack~\cite{tramer2020adaptive}. Taking the stochastic defense methods into account, the BPDA+EOT attack has become the default one when evaluating the state-of-the-art adversarial purification methods~\cite{hill2021stochastic,yoon2021adversarial}. To this end, we use the BPDA+EOT implementation of \cite{hill2021stochastic}: \url{https://github.com/point0bar1/ebm-defense} with default hyperparameters for evaluation.

	\subsection{Implementation details of baselines}
	\label{sec:basline_supp}
	
	\paragraph{Purification models on CelebA-HQ}
	We mainly consider the state-of-the-art VAEs and GANs as purification models for comparison, and in particular we use NVAE~\cite{vahdat2020NVAE} and StyleGAN2~\cite{karras2020analyzing} in our experiments. 
	To use NVAE as a purification model, we directly pass the adversarial images to its encoder and get the purified images from its decoder. To use StyleGAN2 as a purification model, we consider three GAN inversion methods to first invert adversarial images into the latent space of StyleGAN2, and then get the purified images through the StyleGAN2 generator. 
	The three GAN inversion methods that we use in our experiments are as follows: 
	\begin{itemize}
		\item GAN+\textsc{Opt}, an optimization-based GAN inversion method that minimizes the perceptual distance between the output image and the input image w.r.t the w+ latent code. We use the codebase: \url{https://github.com/rosinality/stylegan2-pytorch/blob/master/projector.py} that closely follows the idea of \cite{karras2020analyzing} for the GAN+\textsc{Opt} implementation. The only difference is that the number of optimization iterations we use is $n=500$ to save the computational time while the original number of optimization iterations is $n=1000$. We find that for $n \geq 500$, the recovered images of GAN+\textsc{Opt} do not change much if we increases $n$. 
		\item GAN+\textsc{Enc}, an encoder-based GAN inversion method that uses an extra encoder to encode the input image to the w+ latent code. We use the codebase: \url{https://github.com/eladrich/pixel2style2pixel} corresponding to the idea of pixel2style2pixel (pSp)~\cite{richardson2021encoding} for the GAN+\textsc{Enc} implementation.
		\item GAN+\textsc{Enc}+\textsc{Opt}, which combines the optimization-based and encoder-based GAN inversion methods. We use the codebase: \url{https://github.com/chail/gan-ensembling} corresponding to the idea of \cite{chai2021ensembling} that uses the w+ latent code from the encoder as the initial point for the optimization with $n=500$ iterations.
	\end{itemize}


	\subsection{Other sampling strategies}
	\label{sec:other_sampling}
	
	Here we provide more details of other sampling strategies based on the same pre-trained diffusion models. 
	
	\paragraph{LD-SDE}
	We denote an adversarial image by $\x_a$ and the corresponding clean image by $\x$. Currently with the Langevin dynamics (LD) sampling for purification, given $\x_a$ we are searching for $\x$ freely~\cite{hill2021stochastic,yoon2021adversarial}. Our approach can be considered as conditional sampling of clean image given the adversarial image using $p(\x|\x_a) = \int p(\x(t^*)| \x_a) p(\x|\x(t^*)) d\x(t^*)$ where $p(\x(t^*)| \x_a)$ is first sampled by following the forward diffusion and $p(\x|\x(t^*))$ is then sampled by following the reverse diffusion process. 
	However, there are different ways in which one can formulate this conditional sampling without introducing forward and reverse diffusion processes.
	
	We can write $p(\x|\x_a) \propto p(\x) p(\x_a|\x)$ where $p(\x)$ represent the distribution of clean images and $p(\x_a|\x)$ is the distribution adversarial image given the clean image which we will approximate it by a simple Gaussian distribution $p(\x_a|\x) = N(\x_a; \x, \sigma^2 \eye_d)$. We also do not have access to the true clean data distribution, but we will assume that $p(\x) \propto e^{h(\x)}$ is denoted by a trained generator (more on this later).
	
	As we can see above, we can assume that $p(\x|\x_a) \propto e^{-E(\x|\x_a)}$ with energy function $E(\x|\x_a ) = -h(\x) + \frac{||\x_a - \x ||_2^2}{2\sigma^2}$. 
	Similarly, sampling from $p(\x|\x_a)$ can be done by running the overdamped LD:
	\begin{align}
	\x_{t+1} = \x_t - \frac{\lambda \Delta_t}{2}\nabla_{\x_t} E(\x_t|\x_a ) + \eta \sqrt{\lambda \Delta_t}\eps_t
	\end{align}
	where $\lambda \Delta_t$ is the learning rate and $\eta$ denotes the damping coefficient. When $\Delta_t$ is infinitely small, it corresponds to solving the following forward SDE (termed LD-SDE):
	\begin{equation}
	d\x = -\frac{\lambda}{2} \nabla_\x E(\x |\x_a) dt + \eta \sqrt{\lambda} d \w
	\end{equation}
	for $t \in [0, 1]$ where $\w$ is the standard wiener process. 
	Note the SDE above does not involve any diffusion and denoising and it only uses LD for sampling from a fixed energy function. Recall that $\nabla_\x h(\x)$ is exactly what we have learned by the score function at timestep $t{=}0$ in diffusion models, \textit{i.e.}, $\nabla_x h(x) \approx \s_\theta(\x, 0)$. Thus, the LD-SDE formulation is
	\begin{align}\label{ld_sde}
	d\x = -\frac{1}{2} \left(-\s_\theta(\x, 0) + \frac{\x - \x_a}{\sigma^2} \right) \lambda  dt + \eta \sqrt{\lambda} d {\w}
	\end{align}
	Note that there exist three hyperparameters $(\sigma^2, \lambda, \eta)$ that we have to tune for the best performance. In particular, $\sigma^2$ controls the balance of the attraction term $\x - \x_a$ (to make $\x$ stay close to $\x_a$) and the score function $-\s_\theta(\x, 0)$ (to make $x$ follow the clean data distribution). When $\sigma^2$ becomes infinitely large, there is no attraction term $\x - \x_a$, and the LD-SDE defined in Eq. (\ref{ld_sde}) reduces to the SDE formulation of the normal LD sampling~\cite{grathwohl2020your,hill2021stochastic,yoon2021adversarial}. Note that with this SDE formulation of the LD sampling, we can use the adjoint method as discussed in Section~\ref{sec:adapt_attack} for an evaluation with strong adaptive attacks, such as AutoAttack.
	
	In experiments, to find the best set of hyperparameters $(\sigma^2, \lambda, \eta)$, we first perform a grid search on $\sigma^2 = \{0.001, 0.01, 0.1, 1, 10, 100\}$, $\lambda = \{0.01, 0.1, 1, 10 \}$, $\eta = \{0.1, 1, 5, 10\}$. We find that the best performing configuration is $\sigma^2=100, \; \lambda=0.1, \; \eta=1.0$. Since $\sigma^2=100$ works the best, it implies that LD-SDE performs better without the attraction term $\x - \x_a$ in Eq. (\ref{ld_sde}).

	\begin{table}[ht] 
		\centering
		\vspace{-5pt}
		\caption{Robust accuracies of baselines obtained from RobustBench vs. from our experiments against $\ell_\infty$ threat model ($\epsilon={8}/{255}$) with AutoAttack on CIFAR-10, obtained by different classifier architectures. 
		}
		\label{tab_L_inf_cifar10_supp}
		\footnotesize\addtolength{\tabcolsep}{-2pt}
		\vspace{3pt}
		\begin{tabular}{cccc}
			\hline
			\multirow{2}{*}{Method} &
			\multirow{2}{*}{Extra Data} & Robust Acc & Robust Acc
			\\
			& &  (from RobustBench) & (from our experiments) \\
			\hline & \\[-2ex]
			\rowcolor{LightCyan}
			WideResNet-28-10 & & &
			\\
			\hline
			\text{ \cite{zhang2020geometry}} & \cmark & 59.64  & 59.96
			\\
			\text{ \cite{wu2020adversarial}} & \cmark & 60.04 & 62.11
			\\
			\text{ \cite{gowal2020uncovering}} & \cmark &  62.80 & 62.70
			\\
			\hline
			\text{ \cite{wu2020adversarial}} & \xmark &  56.17 & 59.18
			\\
			\text{ \cite{rebuffi2021fixing}} & \xmark &  60.75 & 61.72
			\\
			\text{ \cite{gowal2021improving}} & \xmark &  63.44 & 65.24 \\ 
			\hline & \\[-2ex]
			\rowcolor{LightCyan}
			WideResNet-70-16 & & &
			\\
			\hline
			\text{ \cite{gowal2020uncovering}} & \cmark &  65.88 & 66.02
			\\
			\text{ \cite{rebuffi2021fixing}} & \cmark &  66.58 & 68.56
			\\
			\hline
			\text{ \cite{gowal2020uncovering}} & \xmark &  57.20 & 59.57
			\\
			\text{ \cite{rebuffi2021fixing}} & \xmark &  64.25 & 64.46
			\\ 
			\text{ \cite{gowal2021improving}} & \xmark &  66.11 & 66.60
			\\ 
			\hline
		\end{tabular}
		\vspace{-10pt}
	\end{table}

	\paragraph{VP-ODE}
	For the reverse generative VP-SDE defined by:
	\begin{align}
	d\x = -\frac{1}{2}\beta(t)[\x + 2\nabla_\x \log p_t(\x)]dt + \sqrt{\beta(t)} d\bar{\w}
	\end{align}
	where the time flows backward from $t={1}$ to 0 and $\bar{w}$ is the reverse standard Wiener process,
	\citet{song2021score} show that there exists an equivalent ODE whose trajectories share the same marginal probability densities $p_t(x(t))$:
	\begin{align}
	d\x = -\frac{1}{2} \beta(t) \left[ \x + \nabla_\x \log p_t(\x) \right] dt
	\end{align}
	where the idea is to use the Fokker-Planck equation~\citep{sarkka2019applied} to transform an SDE to an ODE (see Appendix D.1 for more details in~\citep{song2021score}).
	Therefore, we can use the above ODE, termed VP-ODE, to replace the reverse generative VP-SDE in our method for purification. Similarly with this VP-ODE sampling, we can also use the adjoint method for an evaluation with strong adaptive attacks, such as AutoAttack.

	\subsection{Gradient computation in an analytic example}
	\label{grad_ana}
	Here we provide a simple example to show that the gradient obtained from the adjoint method will closely match its ground-truth value if the SDE solver has a small numerical error. In this example, we know the analytic solution of gradient through a reverse-time SDE, and thus we can compare the difference between the gradient from solving the augmented SDE in Eq. (\ref{aug_sdeint}) and its analytic solution. 
	
	In specific, we assume the data follows a Gaussian distribution, \textit{i.e.}, $x \sim p_0(x) = \mathcal{N}(\mu_0, \sigma_0^2)$. 
	The nice property about the diffusion process in Eq. (\ref{diffused_sample}) is that if $p_0(x)$ is a Gaussian distribution, $p_{t}(x)$ is also Gaussian for all $t$. 
	From Eq. (\ref{diffused_sample}) in VP-SDE, we have $p_{t}(x) = \mathcal{N}(\mu_{t}, \sigma_{t}^2)$  where $\mu_{t} = \mu_0 \sqrt{\alpha({t})}$ and $\sigma_{t}^2 = 1 - (1 - \sigma_0^2) \alpha({t})$, with $\alpha_t :=  e^{-\int_0^t \beta(s)ds}$. 
	Therefore, if we fix $\mu_0$ and $\sigma_0$ to particular values, we can easily evaluate $p_{t^*}(x)$ at the diffusion timestep $t^*$.
	
	Recall that the reverse process can be described as:
	\begin{align*}
	dx = -\frac{1}{2} \beta(t)[x + 2 \nabla_x \log p_t(x)] dt + \sqrt{\beta(t)} d\bar{w}
	\end{align*}
	where $w(t)$ is a standard reverse-time Wiener process. Since we have $p_t(x)$ analytically, we can write $\nabla_x \log p_t(x) = - \frac{x - \mu_{t}}{\sigma_{t}^2}$. So the reverse process is:
	\begin{align}\label{sdeint_supp}
	{x}(0) = \texttt{sdeint}(x(t^*), f_{\text{rev}}, g_{\text{rev}}, \bar{w}, t^*, 0)
	\end{align}
	where the drift and diffusion coefficients are given by
	\begin{align*}
	\begin{split}
	f_{\text{rev}}(x, t) &:= -\frac{1}{2} \beta(t) \left[\frac{(\sigma_{t}^2 - 2)x + 2\mu_{t}}{\sigma_{t}^2} \right] \\
	g_{\text{rev}}(t) &:= \sqrt{\beta(t)}
	\end{split}
	\end{align*}
	Then, we can compute gradient (denoted by ${\phi}_\text{adj}$) of $x(0)$ w.r.t. $x(t^*)$ through the SDE in Eq. (\ref{sdeint_supp}) using the adjoint method in Eq. (\ref{aug_sdeint}), where we use the objective $\mathcal{L} := x(0)$ for simplicity.
	
	On the other hand, let $x(t):=x_t$, we can evaluate $p(x_0 |x_t)$ as well which is
	\begin{align}\label{eq:denoising}
	\begin{split}
	p(x_0 |x_t) &\propto p(x_0) p(x_t|x_0) \propto \exp{ \left( -\frac{(x_0-\mu_0)^2}{2 \sigma_0^2} -\frac{(x_t - x_0 \sqrt{\alpha_t})^2}{2 (1 - \alpha_t)} \right)} \\
	& \propto \exp{ \left( -\frac{(1-\alpha_t + \sigma_0^2 \alpha_t)x_0^2 - 2((1 - \alpha_t)\mu_0 + \sigma_0^2\sqrt{\alpha_t}x_t) x_0}{2(1-\alpha_t)\sigma_0^2} \right) } \\
	& = \mathcal{N}(\frac{(1 - \alpha_t) \mu_0 + \sigma_0^2\sqrt{\alpha_t}x_t}{1 - \alpha_t + \sigma_0^2 \alpha_t}, \frac{(1-\alpha_t)\sigma_0^2}{1-\alpha_t + \sigma_0^2 \alpha_t}) 
	\end{split}
	\end{align}
	where we use $p(x_t|x_0) = \mathcal{N}(x_0 \sqrt{\alpha_t}, 1 - \alpha_t)$ by following Eq. (\ref{diffused_sample}). Thus, we can get the analytic solution (denoted by ${\phi}_\text{ana}$) of the gradient of $x(0)$ w.r.t. $x(t^*)$ as follows:
	\begin{align}
	{\phi}_\text{ana} := \frac{\partial x(0)}{\partial x(t^*)} = \frac{\sigma_0^2\sqrt{\alpha_{t^*}}}{1 - \alpha_{t^*} + \sigma_0^2 \alpha_{t^*}}
	\end{align}
	
\begin{figure}[h]
		\centering
		\includegraphics[width=0.45\linewidth]{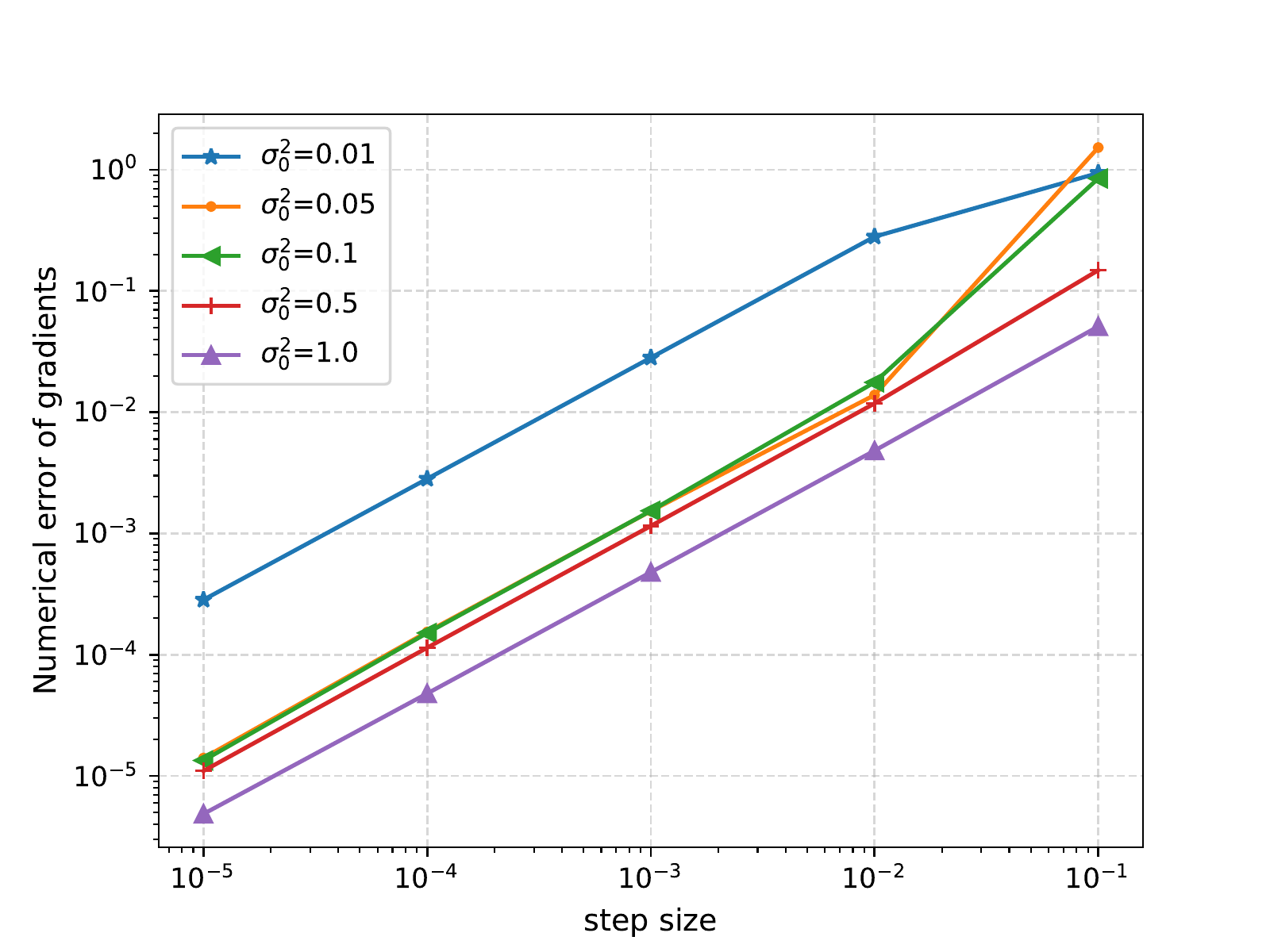}
		\vspace{-8pt}
		\caption{\small Impact of the step size in the Euler-Maruyama method to solve our SDEs on the numeric error of gradients with different $\sigma_0^2$ values. As the step size gets smaller, the numeric error monotonically decreases at the same rate in different settings.}
		\vspace{-5pt}
		\label{impact_step_size}
	\end{figure}

	In experiments, we set $\mu_0=0$ and $\sigma_0^2 \in \{0.01, 0.05, 0.1, 0.5, 1.0\}$, and we use the Euler-Maruyama method to solve our SDEs with different scales of step sizes. The difference between the numeric gradient and the analytic gradient vs. the step size is shown in Figure \ref{impact_step_size}, where the numeric error of gradients is measured by $|{\phi}_\text{ana} - {\phi}_\text{adj}| / |{\phi}_\text{ana}|$. 
	As the step size gets smaller, the numeric error monotonically decreases at the same rate in different settings. It implies that the gradient obtained from the adjoint method will closely match its ground-truth value if the step size in the SDE solver is small.

	\section{More experimental results}

	\subsection{Robust accuracies of our method for standard attack and black-box attack}
	\label{app_sec_stand_black}
	
	In general adaptive attacks are considered to be stronger than standard attack (\textit{i.e.}, non-adaptive). Following the checklist of \citet{croce2022evaluating}, we report the performance of DiffPure for standard attacks in Table~\ref{app_tab_standard}.
	We can see that 1) AutoAttack is effective on the static model as its robust accuracies are zero, and 2) standard attacks are not effective on our method as our robust accuracies against standard attacks are much better than those against adaptive attacks (\textit{ref.} Tables \ref{tab_L_inf_cifar10}-\ref{tab_imagenet}).

	The AutoAttack (the ``standard'' version) we have considered includes the black-box Square Attack, but evaluating with Square Attack separately is a more direct way to show insensitivity to gradient masking. We thus show the performance of DiffPure for Square Attack in Table~\ref{tab_square}.
	We can see that 1) our method has much higher robust accuracies against Square Attack than the static model, and 2) our robust accuracies against Square Attack are higher than those against AutoAttack (\textit{ref.} Table \ref{tab_L_inf_cifar10}-\ref{tab_imagenet}). These results directly show DiffPure's insensitivity to gradient masking.
	
	\begin{table}[h]
		\footnotesize\addtolength{\tabcolsep}{-2.pt}
		\centering
		\vspace{-5pt}
		\caption{Robust accuracies of our method against standard attacks, where we transfer AutoAttack from the static model (i.e., the classifier) to our method DiffPure, and the other evaluation setting and hyperparameters are the same as those in Tables \ref{tab_L_inf_cifar10}-\ref{tab_imagenet}.
		}
		\label{app_tab_standard}
		\begin{tabular}{cccccc}
			\hline
			Dataset & Network & $\ell_p$-norm & Static Model & Ours \\
			\hline
			CIFAR-10 & WRN-28-10 & $\ell_\infty$ & 0.00 & 89.58$\pm$0.49 \\
			CIFAR-10 & WRN-28-10 & $\ell_2$ & 0.00 & 90.37$\pm$0.24 \\
			ImageNet & ResNet-50 & $\ell_\infty$ & 0.00 & 67.01$\pm$0.97 \\
			\hline
		\end{tabular}
		\vspace{-10pt}
	\end{table}
	
	\begin{table}[h]
		\footnotesize\addtolength{\tabcolsep}{-2.pt}
		\centering
		\vspace{-5pt}
		\caption{Robust accuracies of the static model (i.e., the classifier) and our method DiffPure against Square Attack, where the other evaluation setting and hyperparameters are the same as those in Tables \ref{tab_L_inf_cifar10}-\ref{tab_imagenet}.
		}
		\label{tab_square}
		\begin{tabular}{cccccc}
			\hline
			Dataset & Network & $\ell_p$-norm & Static Model & Ours \\
			\hline
			CIFAR-10 & WRN-28-10 & $\ell_\infty$ & 0.33 & 85.42$\pm$0.65 \\
			CIFAR-10 & WRN-28-10 & $\ell_2$ & 21.42 & 88.02$\pm$0.23 \\
			ImageNet & ResNet-50 & $\ell_\infty$ & 9.25 & 62.88$\pm$0.65 \\
			\hline
		\end{tabular}
		\vspace{-10pt}
	\end{table}

	\begin{table}[ht] 
		\centering
		\vspace{-5pt}
		\caption{Robust accuracies of baselines obtained from RobustBench vs. from our experiments against $\ell_2$ threat model ($\epsilon=0.5$) with AutoAttack on CIFAR-10, obtained by different classifier architectures. 
			Methods marked by $^*$ use WideResNet-34-10, with the same width but more layers than the default one.}
		\label{tab_L2_cifar10_supp}
		\footnotesize\addtolength{\tabcolsep}{-2pt}
		\vspace{3pt}
		\begin{tabular}{cccc}
			\hline
			\multirow{2}{*}{Method} &
			\multirow{2}{*}{Extra Data} & Robust Acc & Robust Acc
			\\
			& &  (from RobustBench) & (from our experiments) \\
			\hline & \\[-2ex]
			\rowcolor{LightCyan}
			WideResNet-28-10 & & &
			\\ \hline
			\text{ \cite{augustin2020adversarial}}$^*$ & \cmark & 76.25 & 77.93
			\\ \hline
			\text{ \cite{rony2019decoupling}} & \xmark &  66.44 & 66.41 \\ 
			\text{ \cite{ding2020mma}} & \xmark &  66.09 & 67.77 \\ 
			\text{ \cite{wu2020adversarial}}$^*$ & \xmark &  73.66 & 72.85
			\\
			\text{ \cite{sehwag2021robust}}$^*$ & \xmark &  76.12 & 75.39
			\\
			\text{ \cite{rebuffi2021fixing}} & \xmark & {78.80} & 78.32 \\ 
			\hline & \\[-2ex]
			\rowcolor{LightCyan}
			WideResNet-70-16 & & & \\ \hline
			\text{ \cite{gowal2020uncovering}} & \cmark &  80.53 & 79.88
			\\
			\text{ \cite{rebuffi2021fixing}} & \cmark &  82.32 & 81.44 
			\\ 
			\hline
			\text{ \cite{gowal2020uncovering}} & \xmark &  74.50 & 74.03
			\\
			\text{ \cite{rebuffi2021fixing}} & \xmark &  {80.42} & 80.86 \\ 
			\hline
		\end{tabular}
	\end{table}

	\begin{table}[ht] 
		\centering
		\vspace{-5pt}
		\caption{Robust accuracies of baselines obtained from RobustBench vs. from our experiments against $\ell_\infty$ threat model ($\epsilon={4}/{255}$) with AutoAttack on ImageNet, obtained by different classifier architectures. 
		}
		\label{tab_imagenet_supp}
		\footnotesize\addtolength{\tabcolsep}{-2pt}
		\vspace{3pt}
		\begin{tabular}{cccc}
			\hline
			\multirow{2}{*}{Method} &
			\multirow{2}{*}{Extra Data} & Robust Acc & Robust Acc
			\\
			& &  (from RobustBench) & (from our experiments) \\
			\hline & \\[-2ex]
			\rowcolor{LightCyan}
			ResNet-50 & & & \\ \hline
			\text{ \cite{robustness}} & \xmark &  29.22 & 31.06
			\\
			\text{ \cite{wong2020fast}} & \xmark &  26.24 & 26.95
			\\
			\text{ \cite{salman2020adversarially}} & \xmark  & 34.96 & 37.89
			\\
			\hline & \\[-2ex]
			\rowcolor{LightCyan}
			WideResNet-50-2 & & & \\ \hline
			\text{ \cite{salman2020adversarially}} & \xmark  & 38.14 & 39.25
			\\
			\hline
		\end{tabular}
		\vspace{-5pt}
	\end{table}

	\subsection{Robust accuracies of baselines obtained from RobustBench vs. from our experiments}
	\label{rob_acc_from_ours_vs_}
	
	When we compare with the state-of-the-art adversarial training methods in the RobustBench benchmark, we use the  default hyperparameters for the AutoAttack evaluation. However, since the computational time of evaluating our method with AutoAttack is high (usually taking 50-150 number of function evaluations per attack iteration), we compare the robust accuracy of our method with baselines on a fixed subset of 512 images that is randomly sampled from the test set. 
	
	To show the validity of the results on this subset, we compare the robust accuracies of baselines reported from RobustBench (on the whole test set) vs. from our experiments (on the sampled subset), shown in Tables~\ref{tab_L_inf_cifar10_supp}-\ref{tab_imagenet_supp}. 
	We can see that for different datasets (CIFAR-10 and ImageNet) and network architectures (ResNet and WideResNet), the gap in robust accuracies of most baselines is small (\textit{i.e.}, less than 1.5\% discrepancy). Furthermore, the relative performances of different methods remain the same. These results demonstrate that it is both efficient and effective to evaluate on the fixed subset.

	\subsection{More results of comparison within adversarial purification on CelebA-HQ}
	\label{sec:comp_celeba_supp}

	Here we compare with other adversarial purification methods by using the BPDA+EOT attack with $\ell_\infty$ perturbations on the \textit{smiling} attribute classifier for CelebA-HQ.
	The results are shown in Table~\ref{tab_celeba_purify_supp}.
	We can see that our method still largely outperforms all the baselines, with an absolute improvement of at least +18.78\% in robust accuracy. Compared with the \textit{eyeglasses} attribute, the \textit{smiling} attribute is more difficult to classify, posing a bigger challenge to the defense method. Thus, the robust accuracies of most defense methods are much worse than those with the \textit{eyeglasses} attribute classifier.
	
	\begin{table}[ht]
		\footnotesize\addtolength{\tabcolsep}{-2pt}
		\centering
		\vspace{-5pt}
		\caption{We evaluate with the BPDA+EOT attack on the \textit{smiling} attribute classifier for CelebA-HQ, where $\epsilon=16/255$ for the $\ell_\infty$ perturbations. Note that \textsc{Opt} and \textsc{Enc} denote the optimization-based and econder-based GAN inversions, respectively, and \textsc{Enc}+\textsc{Opt} implies a combination of \textsc{Opt} and \textsc{Enc}.}
		\label{tab_celeba_purify_supp}
		\begin{tabular}{ccccc}
			\hline
			Method & Purification & Standard Acc & Robust Acc \\
			\hline
			\cite{vahdat2020NVAE} & VAE & 93.55 & 0.00 \\
			\cite{karras2020analyzing} & GAN+\textsc{Opt} & 93.49 & 3.41 \\
			\cite{chai2021ensembling} & GAN+\textsc{Enc}+\textsc{Opt} & \textbf{93.68} & 0.78 \\
			\cite{richardson2021encoding} & GAN+\textsc{Enc} & 90.55 & 40.40 \\
			\hline
			Ours ($t^* = 0.4$) & Diffusion & 89.78$\pm$0.14 & 55.73$\pm$0.97 \\
			Ours ($t^* = 0.5$) & Diffusion & 87.62$\pm$0.22 & \textbf{59.12$\pm$0.37}
			\\
			\hline
		\end{tabular}
		\vspace{-5pt}
	\end{table}

	\subsection{More results of ablation studies}
	\label{sec:abalation_supp}

\begin{figure}[ht]
		\centering
		\vspace{-5pt}
		\includegraphics[width=0.35\linewidth]{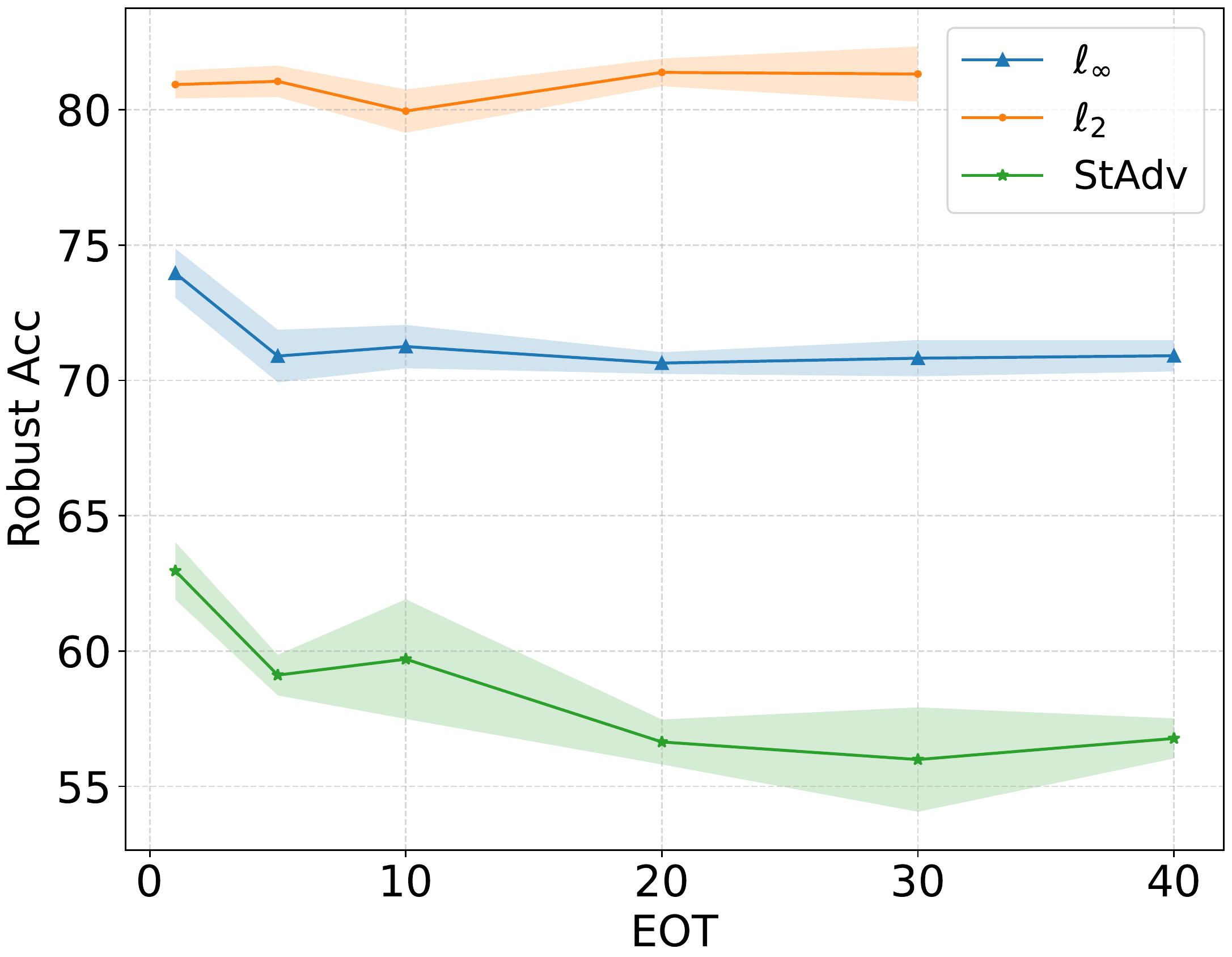}
		\caption{\small Impact of EOT on robust accuracies against the $\ell_\infty$ ($\epsilon=8/255$), $\ell_2$ ($\epsilon=0.5$) and StAdv ($\epsilon=0.05$) threat models, respectively, where we evaluate on WideResNet-28-10 for CIFAR-10.}
		\label{eot}
		\vspace{-5pt}
		\label{impact_eot}
	\end{figure}

	\paragraph{Impact of EOT}
	Because of the stochasticity in our purification process, we seek for the EOT value that is sufficient for different threat models to evaluate our method. In Figure \ref{eot}, we present the robust accuracies of our method over three threat models - $\ell_\infty$, $\ell_2$ and StAdv, respectively, with different number of EOT. 
	We can see that these threat model has different behaviors with the number of EOT: our robust accuracy against the $\ell_2$ 
	threat model seems to be not affected by EOT while our robust accuracies against the $\ell_\infty$ 
	and StAdv threat models first decrease and then saturate as the number of EOT increases. 
	In particular, our robust accuracy saturates at EOT=5 for $\ell_\infty$ and at EOT=20 for StAdv. Therefore, we consider EOT=20 by default for all our experiments unless stated otherwise, which should be sufficient for threat models to evaluate our method.
	
	\paragraph{Randomizing diffusion timestep $t^*$} 
	Since randomness matters for adversarial purification methods~\citep{hill2021stochastic,yoon2021adversarial}, we here consider introduce another source of randomness by randomizing the diffusion timestep $t^*$: 
	Instead of using a fixed $t^*$, we uniformly sample $t^*$ from the range $[\bar{t} - \Delta_t, \bar{t} + \Delta_t]$ for every diffusion process. Table \ref{tab_rand_t} shows the robustness performance with different $\Delta_t$, where a larger $\Delta_t$ means stronger randomness introduced by perturbing $t^*$.
	We can see that the mean of standard accuracy monotonically decreases and the variance of robust accuracy monotonically increases with $\Delta_t$ due to the stronger randomness. Besides, a slightly small $\Delta_t$ may improve the robust accuracy in an average sense, while a large $\Delta_t$ may hurt it. It implies that there may also exist a sweet spot for the perturbation strength $\Delta_t$ of diffusion timestep $t^*$ to get the best robustness performance.

	\begin{table}[ht] 
		\vspace{-8pt}
		\centering
		\caption{Uniformly sampling diffusion timestep $t^*$ from $[\bar{t} - \Delta_t, \bar{t} + \Delta_t]$, where $\bar{t}=0.1$. We evaluate with AutoAttack (the \textsc{Rand} version with EOT=20) $\ell_\infty$ ($\epsilon=8/255$) on WideResNet-28-10 for CIFAR-10. Note that $\Delta_t=0$ reduces to our method with a fixed $t^*$.}
		\label{tab_rand_t}
		\footnotesize\addtolength{\tabcolsep}{0pt}
		\vspace{3pt}
		\begin{tabular}{cccc}
			\hline
			$\Delta_t$ & Standard Acc & Robust Acc \\
			\hline
			0 & \textbf{89.02$\pm$0.21} & 70.64$\pm$0.39 \\
			0.015 & 88.86$\pm$0.22 & \textbf{72.14$\pm$1.45} \\
			0.025 & 88.04$\pm$0.33 & {69.21$\pm$2.74}
			\\
			\hline
		\end{tabular}
	\end{table}

	\subsection{Purifying adversarial examples of standard attribute classifiers}
	\label{sup:sec_purify_vis}
	\vspace{-2pt}
	
\begin{figure}[!ht]
		\centering
		\begin{subfigure}{\linewidth}
			\centering
			\includegraphics[width=0.76\linewidth]{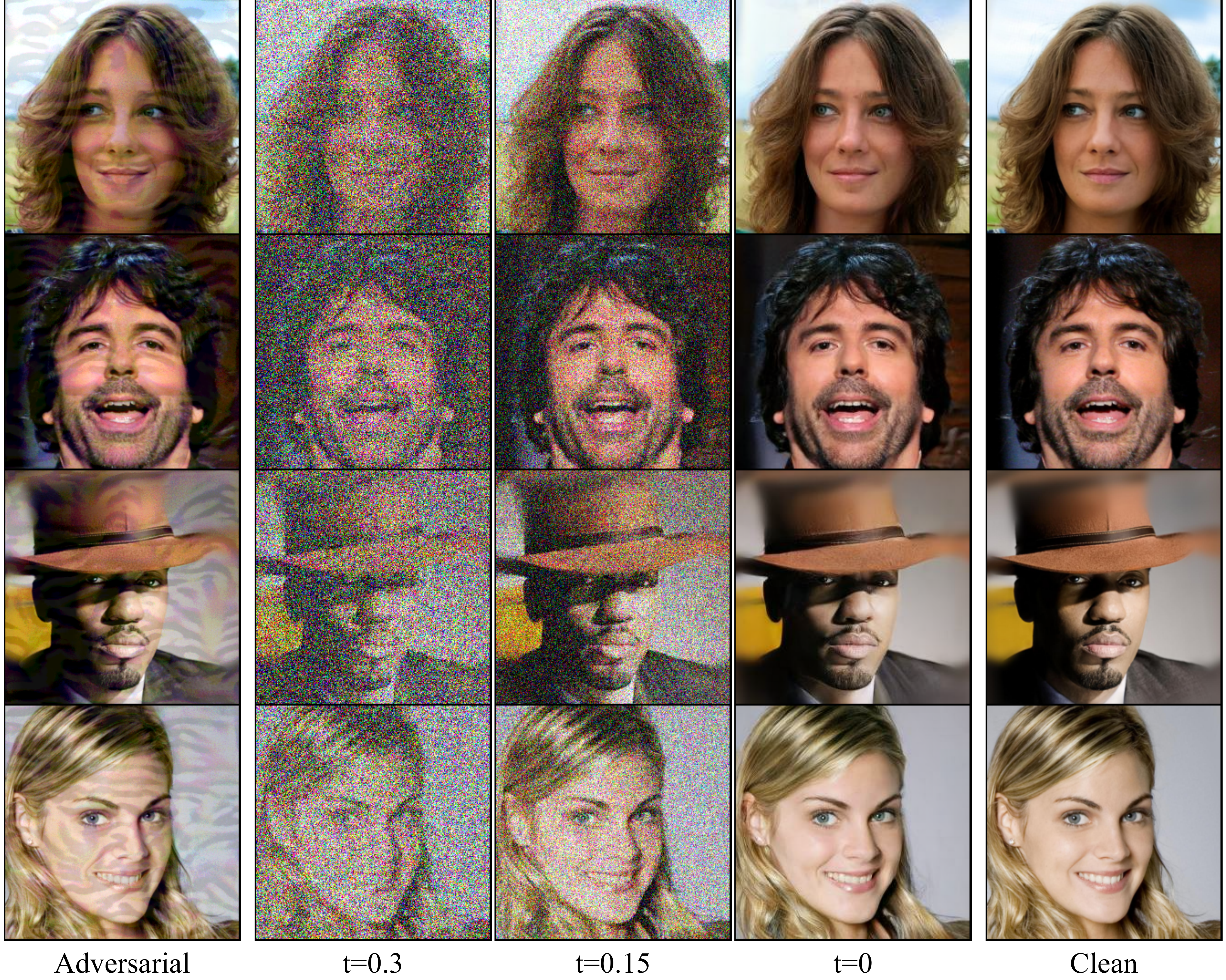}
			\vspace{-3pt}
			\caption{ Smiling}
			\label{smile_supp}
		\end{subfigure}
		\begin{subfigure}{\linewidth}
			\centering
			\includegraphics[width=0.72\linewidth]{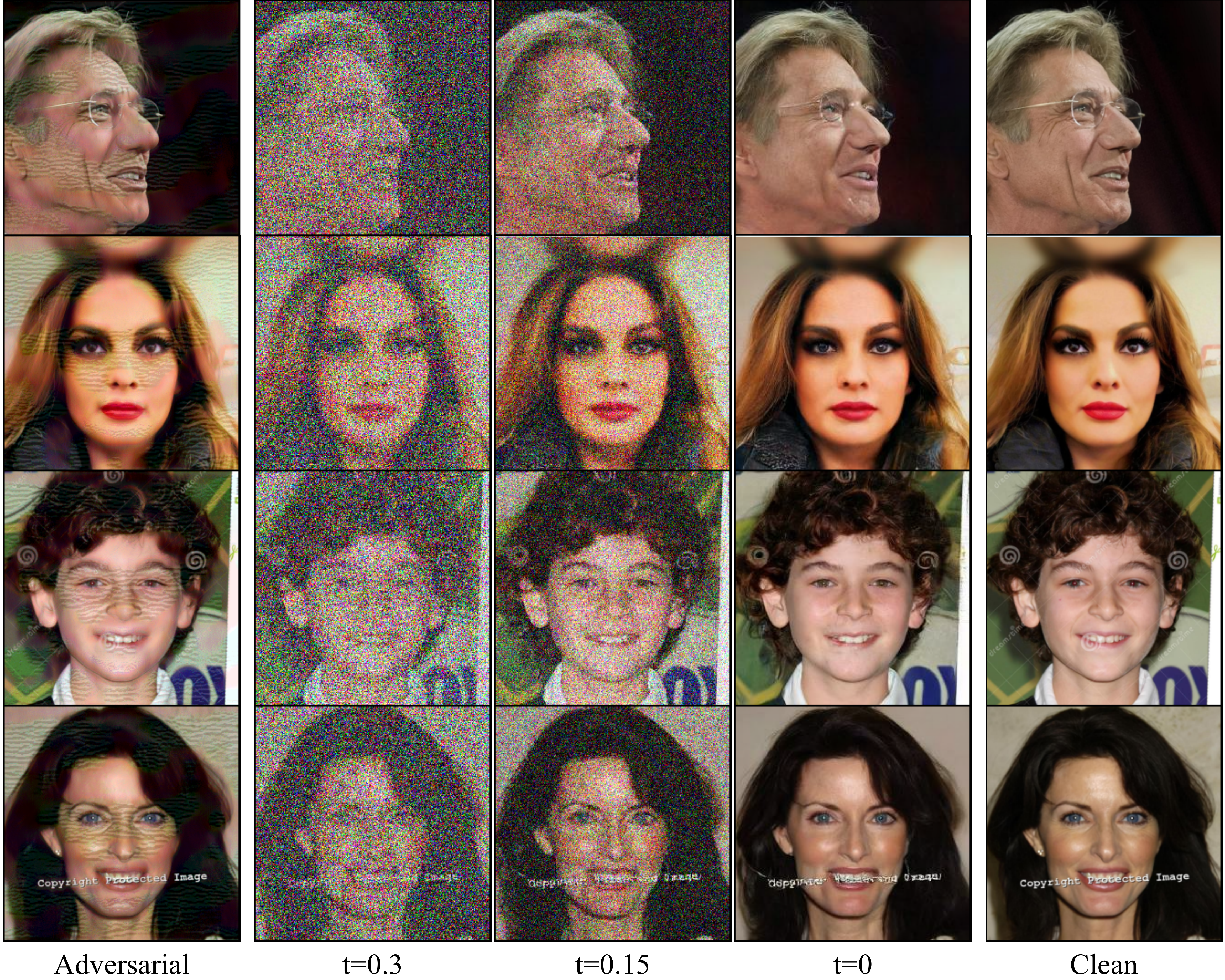}
			\vspace{-3pt}
			\caption{ Eyeglasses}
			\vspace{-8pt}
			\label{eyeglasses_supp}
		\end{subfigure}
		\caption{Our method purifies adversarial examples (first column) produced by attacking attribute classifiers using PGD $\ell_\infty$ ($\epsilon=16/255$), where $t^*=0.3$. The middle three columns show the results of the SDE in Eq. (\ref{sdeint}) at different timesteps, and we observe the purified images at $t{=}0$ match the clean images (last column). Better zoom in to see how we remove adversarial perturbations.}
		\label{vis_adv_supp}
	\end{figure}

	In Figure~\ref{vis_adv_supp}, we provide more visual examples of how our methods purify the adversarial examples of standard classifiers.
	
	\subsection{Inference time with and without DiffPure}
	\label{app_infer_time}
	
	Inference time (in seconds) by varying diffusion timestep $t^*$ is reported in Table~\ref{tab_infer_time}, where the inference time increases linearly with $t^*$. 
	We believe our inference time can be reduced using recent fast sampling methods for diffusion models, but we leave it as the future work.
	
	\begin{table}[h]
		\footnotesize\addtolength{\tabcolsep}{-2pt}
		\centering
		\vspace{-5pt}
		\caption{Inference time with DiffPure ($t^*>0$) and without DiffPure ($t^*=0$) for a single image on an NVIDIA V100 GPU, where the time increase over $t^*=0$ is given in parenthesis.
		}
		\label{tab_infer_time}
		\begin{tabular}{ccccccc}
			\hline
			Dataset & Network & $t^*$=0 & $t^*$=0.05 & $t^*$=0.1 & $t^*$=0.15 \\
			\hline
			CIFAR-10 & WRN-28-10 & 0.055 & 5.12\textbf{\tiny ($\times$93)} & 10.56\textbf{\tiny ($\times$190)} & 15.36\textbf{\tiny ($\times$278)} \\
			ImageNet & ResNet-50 & 0.062 & 5.58\textbf{\tiny ($\times$90)} & 11.13\textbf{\tiny ($\times$179)} & 17.14\textbf{\tiny ($\times$276)} \\
			\hline
		\end{tabular}
		\vspace{-5pt}
	\end{table}

	\subsection{Crafting examples just for diffusion model }
	\label{app_sec_attack_diff_model}
	
	It is interesting to see if the adversary can craft examples just for the diffusion model, such that the recovered images from DiffPure become different from the original clean images, which may also result in the misclassification. To this end, we use APGD (EOT=20) to attack the diffusion model only by maximizing the mean squared error (MSE) between diffusion model's outputs and input images. The results are given in Table \ref{tab_diffusion}. We see that attacking the diffusion model is less effective than attacking the whole defense system.

	\begin{table}[!h]
		\footnotesize\addtolength{\tabcolsep}{-2pt}
		\centering
		\vspace{-10pt}
		\caption{Robust accuracies of attacking the diffusion model only (``Diffusion only'') and the whole defense system (``Diffusion+Clf'') using APGD $\ell_\infty$ and  $\ell_2$ on CIFAR-10.
		}
		\label{tab_diffusion}
		\begin{tabular}{ccccc}
			\hline
			$\ell_p$-norm & Network & $t^*$ & Diffusion only & Diffusion+Clf \\
			\hline
			$\ell_\infty$ ($\epsilon$=8/255) & WRN-28-10 & $0.1$ & 85.04$\pm$0.86 & 75.91$\pm$0.74 \\
			$\ell_2$ ($\epsilon$=0.5) & WRN-28-10 & $0.075$ & 90.82$\pm$0.42 & 84.83$\pm$0.09 \\
			\hline
		\end{tabular}
		\vspace{-10pt}
	\end{table}


\end{document}